\documentclass[12pt]{article}
\usepackage{latex-macros}
\usepackage{mathtools}
\usepackage{multirow}
\usepackage[scale=0.75]{geometry}
\usepackage{booktabs}
\usepackage[capitalize]{cleveref}
\usepackage{amssymb,amsthm}
\usepackage{algpseudocode}
\usepackage{appendix}
\usepackage{cancel}
\usepackage{diagbox}
\usepackage{enumitem}
\setlist[enumerate]{leftmargin=.5in}
\setlist[itemize]{leftmargin=.5in}
\usepackage{algorithm}
\usepackage{algpseudocode}
\usepackage[labelfont=bf]{caption}
\usepackage[sort,numbers]{natbib}
\usepackage{subcaption}
\usepackage{xfrac}
\usepackage{xcolor}
\definecolor{lblue}{HTML}{908cc0}
\definecolor{mblue}{HTML}{519cc8}
\definecolor{hblue}{HTML}{1d5996}
\definecolor{lred}{HTML}{cb5501}
\definecolor{mred}{HTML}{f1885b}
\definecolor{hred}{HTML}{b3001e}
\definecolor{ttred}{HTML}{ca3542}

\usepackage{tikz,pgfplots}
\pgfplotsset{compat=1.14}
\usetikzlibrary{automata,decorations.markings,positioning}

\newcommand{\bignorm}[1]{\bigg\| {#1} \bigg\|}

\crefname{appsec}{Appendix}{Appendices}

\newtheorem{assumption}{Assumption}
\Crefname{assumption}{Assumption}{Assumptions}
\Crefname{theorem}{Theorem}{Theorems}

\newtheorem{theorem}{Theorem}
\newtheorem{lemma}{Lemma}

\newtheorem{proposition}{Proposition}

\newcommand{\tsup}[1]{\textsuperscript{#1}}

\newcommand{\est}{\hat{X}}
\newcommand{\mtrv}{\widehat{V}}
\newcommand{\mtrz}{\widehat{Z}}
\newcommand{\mtry}{\widehat{Y}}

\newcommand{\err}{\widehat{E}}
\newcommand{\iid}{\overset{\mathrm{i.i.d.}}{\sim}}

\usepackage{hyperref}
\definecolor{mylinkcolor}{RGB}{0,0,140}
\hypersetup{colorlinks,allcolors=mylinkcolor,citecolor=mylinkcolor}

\title{{Communication-efficient distributed eigenspace estimation}}
\author{Vasileios Charisopoulos\thanks{
Department of Operations Research \& Information Engineering,
Cornell University, 14853 Ithaca NY - email: \url{vc333@cornell.edu}}
\and
Austin R. Benson\thanks{
Department of Computer Science, Cornell University,
14853 Ithaca, NY - email: \url{arb@cs.cornell.edu}}
\and
Anil Damle\thanks{
Department of Computer Science, Cornell University,
14853 Ithaca, NY - email: \url{damle@cornell.edu}}
}

\allowdisplaybreaks

\newcommand{\newstuff}[1]{#1}
\theoremstyle{remark}
\newtheorem{remark}{Remark}

\begin{document}

\maketitle

\begin{abstract}
  Distributed computing is a standard way to scale up machine learning and data science
  algorithms to process large amounts of data.
  In such settings, avoiding communication amongst machines is paramount for
  achieving high performance.
  Rather than distribute the computation of existing algorithms, a common practice
  for avoiding communication is to compute local solutions or parameter estimates on each machine
  and then combine the results; in many convex optimization problems, even simple averaging
  of local solutions can work well.
  However, these schemes do not work when the local solutions are not unique.
  \textit{Spectral methods} are a collection of such problems, where
  solutions are orthonormal bases of the leading invariant subspace of an associated data
  matrix, which are only unique up to rotation and reflections.
  Here, we develop a communication-efficient distributed algorithm for computing the leading invariant
  subspace of a data matrix.
  Our algorithm uses a novel alignment scheme that minimizes the
  Procrustean distance between local solutions and a reference solution,
  and only requires a single round of communication.
  For the important case of principal component analysis (PCA), we show that our algorithm
  achieves a similar error rate to that of a centralized estimator.
  We present numerical experiments demonstrating the efficacy of our proposed
  algorithm for distributed PCA, as well as other problems where solutions
  exhibit rotational symmetry, such as node embeddings for graph data and
  spectral initialization for quadratic sensing.
\end{abstract}

\section{Overview \& Background}
The paradigm of distributed computing, where data collection and/or computation
for a fixed task happens on several interconnected machines, is by now standard
in machine learning and data science~\cite{tensorflow}. Typically, each machine holds its own data
points or samples and a global solution is approximated using only local
computation and communication between machines.
Since communication is the bottleneck operation~\cite{BBL11,ZCL+20}, it is
highly desirable to avoid multiple rounds of communication (i.e. multiple
instances where machines broadcast or exchange information with each other).
One of several flavors of distributed computing, which is the focus of this
paper, is \textit{federated learning}~\cite{KMR15,KMRR16}.
In federated learning, local compute nodes
communicate local information to a central `coordinator', though there are
several other configurations used in distributed computing.
When the underlying problem is ``simple'' (e.g., convex), combining local
information and solutions is relatively well-studied, and even the simplest
schemes (such as one-shot averaging of local solutions) often work well
with minimal communication costs~\cite{ZDM13}.
However, many problems are not amenable to such schemes.

Consider the model problem of distributed computation of the first principal
component in principal component analysis (PCA). We assume that each of $m$ machines
draws $n$ i.i.d.\ samples from some underlying distribution $\cD$ and that, for simplicity,
$\expec[x \sim \cD]{x} = 0$. Then we want to
to estimate the leading eigenvector of the covariance matrix
$\Sigma := \expec[x \sim \cD]{xx^{\T}}$.
Here, solutions are all subject to a natural \textit{sign ambiguity} (in
addition to a scale ambiguity, which is trivial to handle by normalizing); if
$\hat{v}_1^{i}$ is the estimate produced by machine $i$, then $-\hat{v}_1^{i}$
is also a valid local solution. Thus, naively averaging two solutions from two
machines could result in an estimate close to zero. \newstuff{Even though the
resulting estimate may still be aligned with, e.g., $v_1$, its magnitude will
decrease at the same rate as the magnitude of the noise, and thus averaging
offers no improvement.}
Continuing this, suppose we fix some unit-norm eigenvector $v_1$ and assume that
roughly half of the local solutions are aligned with $v_1$ while the other
half are aligned with $-v_1$; it is clear that we should not expect naive
averaging to work in this situation. Indeed, Garber et al.~\cite{GSS17} showed that
the resulting estimate will have $\Omega(\sqrt{\sfrac{1}{n}})$ error.
On the other hand, a centralized estimator with access to all $m \cdot n$ samples
would achieve an error of $\cO(\sqrt{\sfrac{1}{mn}})$.
To address this, they developed a ``sign-fixing'' scheme for combining eigenvectors
to achieve an error rate similar to the centralized estimator.

\begin{figure}[h]
    \centering
    \begin{subfigure}{0.45\textwidth}
        \includegraphics[width=\linewidth]{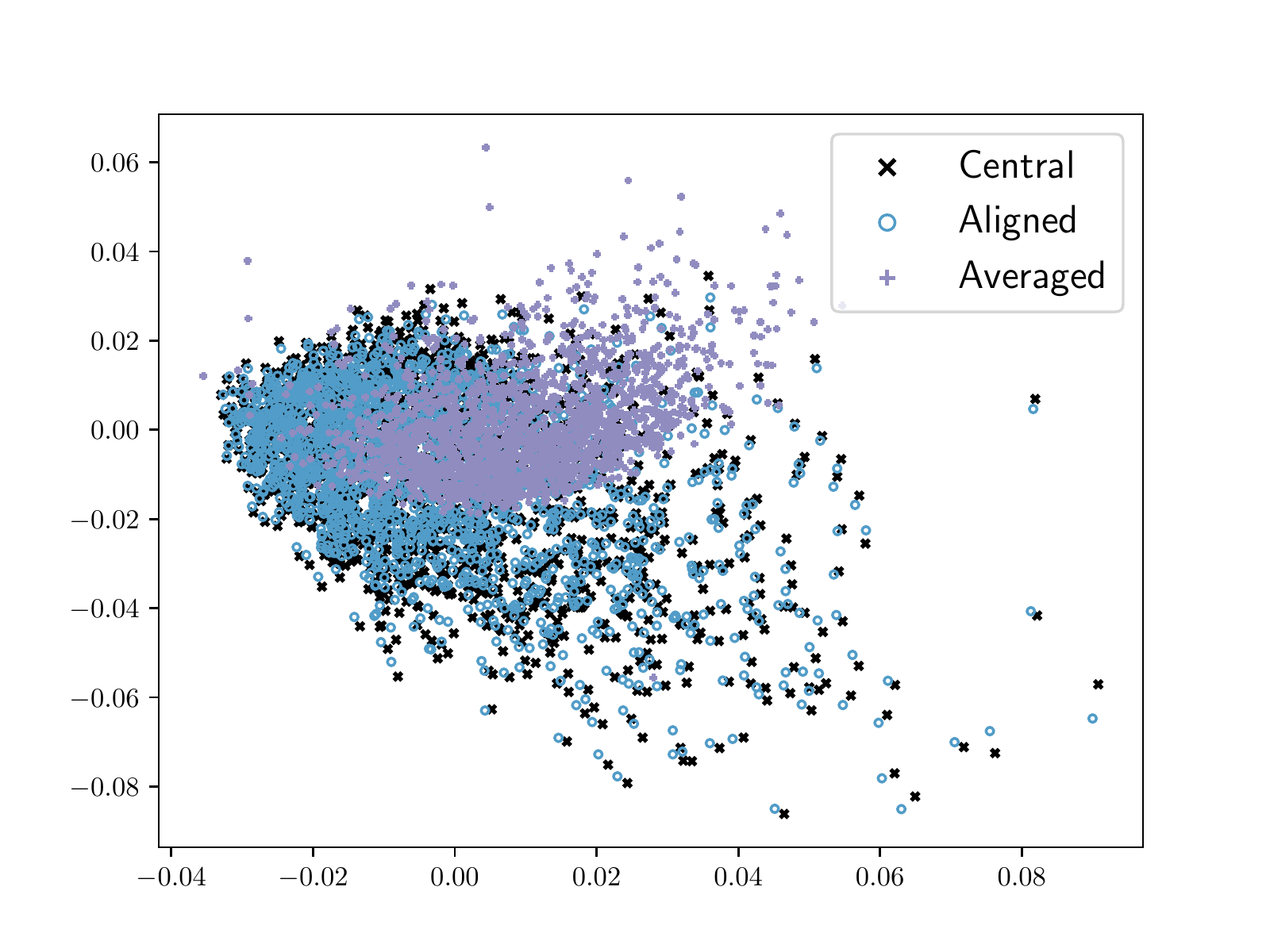}
        \caption{Samples from digit $4$}
    \end{subfigure}
    \begin{subfigure}{0.45\textwidth}
        \includegraphics[width=\linewidth]{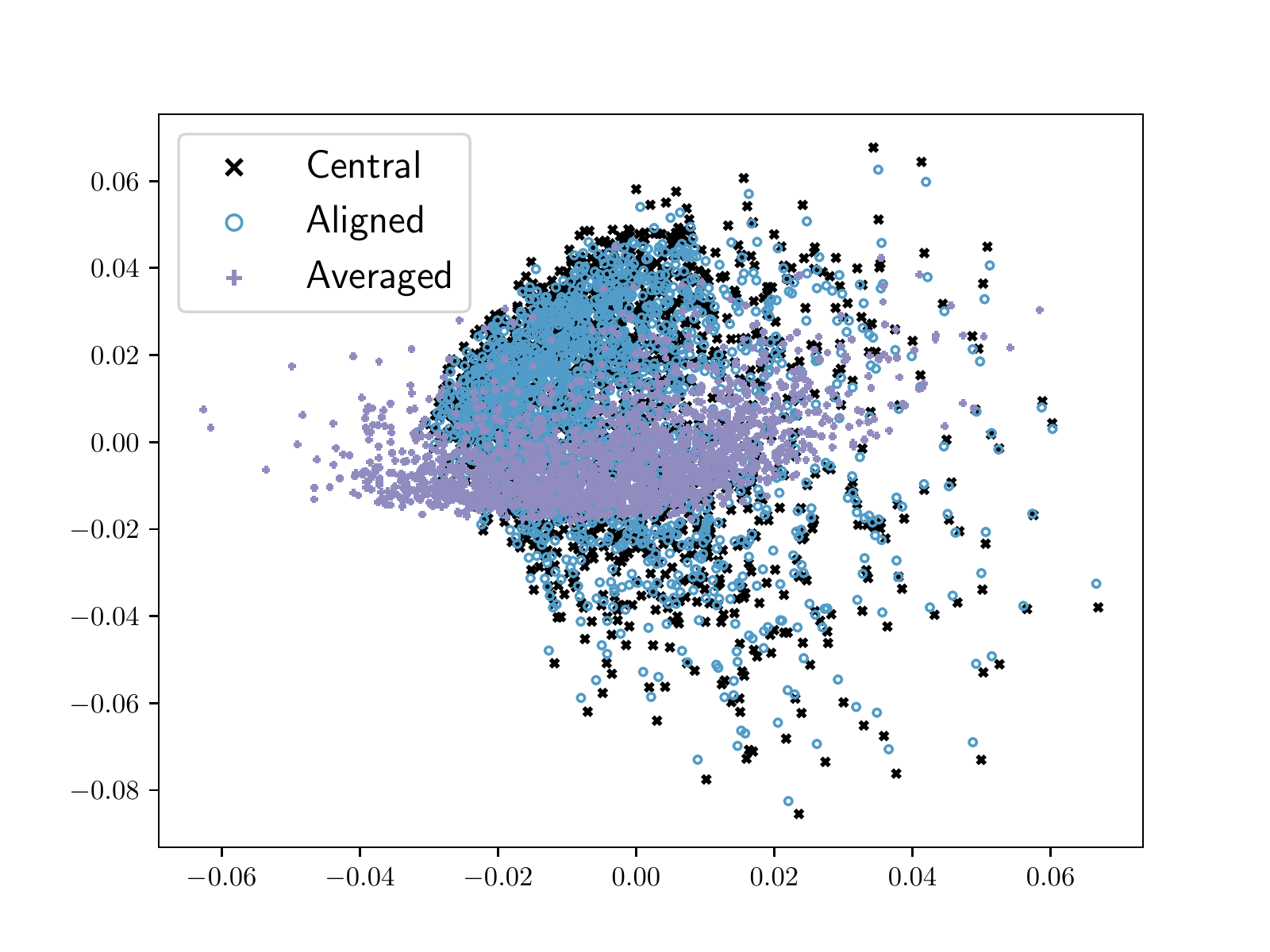}
        \caption{Samples from digit $8$}
    \end{subfigure}
    \caption{Projection of samples to the first two
        principal components of the \texttt{MNIST} dataset, in a distributed setting
        where samples are split across $m = 25$ machines. The central algorithm
        (\textsf{Central}) produces a very different scatterplot than naive
        averaging (\textsf{Average}) of local solutions. In constrast, our
        alignment algorithm (\textsf{Aligned}) leads to a projection very similar
        to that of the central algorithm. The subspace distance of the averaged
        solution is $\approx 0.95$ (indicating that the subspaces are
        near-orthogonal to each other), while the subspace distance of the
        aligned solution is $\approx 0.35$.}
    \label{fig:real-pca}
\end{figure}

When estimating eigenspaces of higher dimension (e.g., the first $r$ principal components for $r > 1$),
we have to deal with an \textit{orthogonal ambiguity} (i.e. for any solution $V$
and orthogonal matrix $Z$, $V Z$ is also a solution), making the task of
combining local solutions highly nontrivial for problems exhibiting natural
symmetries.
In particular, adapting spectral algorithms to the distributed setting poses
a significant challenge, and these algorithms form the basis of a rich set of
applications such as
dimensionality reduction~\cite{Jolliffe02}, clustering~\cite{vonLux07},
ranking~\cite{PBMW99} and high-dimensional estimation~\cite{KV09}.
\Cref{fig:real-pca} depicts the results of distributed PCA on examples from
the \texttt{MNIST} dataset; indeed, naive averaging of local solutions can be
catastrophic in practice. On the other hand, the solution produced by an
appropriate alignment algorithm (in particular, using our Algorithm~\ref{alg:procrustes-fixing})
is very close to that of the central algorithm.

\subsection{Contribution \& outline}
In this paper, we propose a general technique for averaging local
solutions to subspace estimation problems in the distributed setting. In
particular, we assume that every machine $i$ observes a noisy version $\est^i$
of a symmetric ``ground truth'' $X$, and wish to estimate the principal
$r$-dimensional eigenspace of $X$. Our method relies on aligning local
estimates with a reference solution (which can be any of the local solutions),
followed by an averaging step. In the
special case $r = 1$, our method recovers the \textit{sign-fixing} scheme of
\cite{GSS17}.
When the underlying matrix $X$ has a nontrivial eigengap and the local matrices $\est^i$ are not ``too far'' from
$X$, we show that the error of the resulting estimate is a combination of a term
that scales \textit{quadratically} with local errors $E^i := \est^i - X$ and a
term that measures how well the empirical average of the local matrices
approximates $X$. \newstuff{Our main result is a \textbf{deterministic} 
bound that does not require any application-specific information.}

We specialize these results to distributed PCA and show that, with high
probability, the estimate produced by our algorithm matches the error rate of
a centralized estimator, with access to all $m \cdot n$ samples.
In addition, we show that the per-node sample complexity scales with the \textit{stable rank}
$r_{\star} := \mathsf{intdim}(\Sigma)$ of the population covariance matrix
$\Sigma$, which is typically much smaller than both
the ambient dimension $d$ and the algebraic rank $r$, for subgaussian designs.
\newstuff{Our rates improve by factor of $\sqrt{r}$ upon those of~\cite{FWWZ19} for subgaussian
distributions, though the upper bound given in the latter work is for the subspace
distance measured in the Frobenius norm and thus not directly comparable with ours
(on the other hand, applying norm equivalence to our result to obtain a bound on the
Frobenius subspace distance exactly recovers the rate of~\cite{FWWZ19}).}
We also recover the result of~\cite{GSS17} when $r = 1$.
Finally, we conduct several numerical experiments that demonstrate the efficacy of our proposed scheme
on applications covering \textbf{(i)} PCA, \textbf{(ii)} node embeddings for graph data, and
\textbf{(iii)} spectral initializations for quadratic sensing.

\subsection{Related work}

Distributed computing has received widespread attention in recent years, leading
to a number of different methods of aggregating information; these include:
\begin{enumerate}
    \item \label{item:dist-sol-1}
        solving subproblems locally and then forming a central solution by
        averaging all local solutions,
    \item \label{item:dist-sol-2}
        ``distributing'' an iterative procedure across machines, for example
        by communicating first and/or second-order information to the coordinator
        to perform a ``central'' gradient step, and
    \item \label{item:dist-sol-3}
        using gossip algorithms, in which individual machines are allowed to
        communicate with peers (instead of only a central coordinator) and can
        aggregate updates from their neighbors.
\end{enumerate}
In the first of the above settings, the average of local solutions may be computed
only once~\cite{ZDM13} or communicated back to the local machines to repeat
multiple rounds of computation~\cite{MMR+17}. The second and third settings
have received widespread attention from the convex optimization community
\cite{NO09,DAW12,SSZ14,JST+14,MSJ+15,SFM+18}, even allowing
\textit{optimal} algorithms in both centralized and decentralized
settings~\cite{SBB+19,ULGN20}. From a statistical perspective,
other works have shown that distributed algorithms can achieve minimax-optimal
rates for statistical estimation~\cite{ZDJW13,DJWZ14,RN16,JLY19}. Lately, a
recurring theme is also \textit{robustness}, to deal with the possibility that
some computing nodes may be malicious~\cite{FXM14,CSX17}.

\paragraph{Dealing with natural symmetries}
As mentioned above, problems with solutions exhibiting natural symmetries
require more careful analyses that address the symmetry at hand. For
spectral methods, a major portion of the literature has focused on distributed
PCA (and the related task of eigenspace estimation of covariance matrices).
For example, recent work formulates PCA in a streaming setting and adapts techniques
such as gradient descent and variance reduction~\cite{Shamir16a,ZL17a,ZL17b}.
However, the resulting methods are not communication-efficient ---
at least not without nontrivial modifications --- as they need to access
sequences of samples that may be
scattered across machines.
Other works~\cite{GSS17,CLLY20} have adapted the shift-and-invert framework~\cite{GHJ+16}
(which reduces an eigenproblem to approximately solving a sequence of linear systems)
to the distributed setting, though the resulting algorithms still require
multiple communication rounds.

More attractive options in terms of communication cost were proposed
in~\cite{FSS13,KVW14,LBKW14} for PCA and related problems. In these algorithms,
each node performs a local SVD and broadcasts its top $r_1 \geq r$ singular values
and vectors $(\Sigma^i, V^i)$, which act as a summary of the local data,
to a central node. That node then forms the matrix
$Y := \bmx{\Sigma^1 (V^1)^\T & \dots & \Sigma^m (V^m)^\T}$ and computes
its SVD, returning the top $r_2 = r$ right singular vectors, where $r$
is the dimension of the desired subspace.
This procedure can also be augmented by sketching to reduce the communication
cost, and adapted to other settings such as kernel PCA~\cite{BLS+16}.
\newstuff{An alternative approach is the Frequent Directions method~\cite{GLPW16},
which can incrementally update a sketch of a matrix that serves as a low-rank
approximation; in particular, the sketches produced by the method are \textit{mergeable},
making the method amenable to parallelization.}
Since the aforementioned works are primarily interested in getting a high quality
low-rank approximation of the data matrices, approximating the leading invariant
subspace of the population covariance matrix in a distributed fashion was
largely overlooked until recently.
\newstuff{One of the first works in this direction is~\cite{KA09}},
which focuses on leading eigenvector estimation for large matrices by aggregating
the eigenvector of randomly sampled submatrices using an alignment step that
removes sign ambiguity. \newstuff{Removing sign ambiguity from singular vectors
is also discussed in~\cite{BAK08}; therein, the authors utilize dataset information
to determine meaningful signs for the computed singular vectors. However, in
settings where the data are random, the algorithm from \cite{BAK08} can still
result in arbitrarily chosen signs; moreover, it attempts to fix signs of individual
singular vectors rather than arbitrary orthogonal ambiguities;
this makes it unsuitable as a preprocessing step for an averaging method when $r > 1$.}

A similar approach to~\cite{KA09} is adopted in~\cite{GSS17}, who
propose an optimal averaging method for computing the leading eigenvector by
aligning all local estimates with a reference solution (e.g. the estimate of the
first node); in addition to showing that the algorithm essentially matches the
performance of ``centralized'' PCA, the authors in~\cite{GSS17} also show that
``naive'' averaging can produce arbitrarily bad estimates. However, they do not
generalize their technique to more than one eigenvector.
More recent works~\cite{BW19,FWWZ19} show that an aggregation method similar to that
of~\cite{LBKW14} achieves error competitive with a centralized estimator even
for $r > 1$, via careful statistical analyses. 
The resulting algorithm deals with orthogonal ambiguity by averaging the local
\textit{spectral projectors} -- however, this leaves open the question of how
to generalize the methods of~\cite{KA09,GSS17} to higher-dimensional subspaces.

\subsection{Notation}
We denote $\set{1, \dots, n}$ by $[n]$
and write $\ip{x, y} = x^\T y$ for vectors
$x$ and $y$, as well as $\ip{X, Y} = \trace{X^\T Y}$ for compatible matrices $X$ and $Y$.
Given $A \in \Rbb^{d_1 \times d_2}$, we write $A_{i, :} \in \Rbb^{d_1}$ for its $i^{\text{th}}$
row vector and $A_{:, j} \in \Rbb^{d_2}$ for its $j^{\text{th}}$ column vector. We write
$\norm{A}_F$ and $\norm{A}_2$ for the Frobenius and spectral norms of $A$ and
$\norm{A}_{2 \to \infty} := \max_{i \in [d_1]} \norm{A_{i, :}}_2$.
We let $\Sbb^{d - 1} := \set{z \in \Rbb^d \mid \norm{z}_2 = 1}$ denote
the unit sphere in $d$ dimensions, and $\Obb_{d_1, d_2}$ denote the set of $d_1 \times d_2$
matrices with orthonormal columns (i.e., the set of $d_1 \times d_2$ orthogonal matrices),
omitting the second subscript when $d_1 = d_2$.
Throughout, we write $\dist_2(U, V) := \norm{UU^{\T} - VV^{\T}}_2$ for the
distance between the subspaces spanned by the columns of $U$ and $V$.
We use the letters $m$ to denote the number
of machines, $n$ to denote the number of samples drawn per machine, and $d$
to denote the dimension of each sample. Finally, we use $A \lesssim B$ to denote that
$A \leq c \cdot B$ for some constant $c$ independent of $m$, $n$ and $d$.

\section{Distributed eigenspace estimation}
\label{sec:distributed-pca}
We study distributed eigenspace estimation when all machines observe a ``noisy''
version of an underlying symmetric matrix $X$. In particular, we assume that
$X$ admits an eigendecomposition of the form
\begin{equation}
    X = \bmx{V_1 & V_2} \bmx{\Lambda_1 & 0 \\ 0 & \Lambda_2} \bmx{V_1 & V_2}^{\T},
    \label{eq:X-evdecomp}
\end{equation}
where $\Lambda_1 := \diag(\lambda_1, \dots, \lambda_r)$ contains the $r$
largest eigenvalues and $\Lambda_2$ contains the $d - r$ smallest
eigenvalues, ordered algebraically so that $\lambda_1 \geq \lambda_2
\dots \geq \lambda_d$. In our setting, the objective is to estimate the leading
$r$-dimensional invariant subspace $V_1$ given a set of $m$ machines with
noisy observations $\est^i \in \Rbb^{d \times d}$, $i \in [m]$, which are also symmetric.
\newstuff{Though our presentation assumes that we are interested in the leading
    $r$-dimensional subspace, this assumption is without loss of generality; our
    results also apply to, e.g., approximating the $r$-dimensional invariant
    subspace corresponding to the \textit{smallest} eigenvalues, since the latter
    can be turned into the leading eigenspace by an appropriate shift.}

To illustrate, consider the setting of distributed PCA. There, every machine
$i$ draws $n$ i.i.d. samples $x_j^{(i)} \in \Rbb^d, \; j \in [n]$ from a
distribution $\cD$ that we assume is zero-mean for the sake of simplicity,
and forms its ``local'' empirical covariance matrix
\begin{equation}
	\est^i := \frac{1}{n} \sum_{j = 1}^n x_j^{(i)} {x_j^{(i)}}^\T.
    \label{eq:local-covmat}
\end{equation}
Here $\expec{\est^i} = \Sigma := \expec[x \sim \cD]{xx^{\T}}$, where $\cD$ is
the underlying distribution, and the task at hand amounts to estimating the
leading eigenspace of the population covariance matrix; \newstuff{without loss
of generality, \textbf{assume that $\lambda_1(\Sigma) = 1$}}.
Note that by standard concentration arguments~\cite{Wainwright19}, a centralized
version of this problem achieves an error rate of roughly
$\tilde{\cO}\big(\sqrt{\sfrac{1}{mn}}\big)$ (with distance measured in the spectral
norm), so we cannot expect a better rate in the distributed setting.

A natural first approach, inspired from one-shot averaging in convex
optimization~\cite{ZDM13}, has each machine compute a local approximation $\hat{V}_1^i$
and then averages all local solutions centrally, with the hope that
averaging will further ``smooth out'' the errors from local solutions:
\begin{equation}
    \text{form} \quad \bar{V} := \frac{1}{m} \sum_{i=1}^m \hat{V}_1^i, \;
    \quad \text{take $Q$ factor from} \quad \texttt{qr}(\bar{V}).
    \label{eq:basic_agg}
\end{equation}
However, this typically fails due to the inherent orthogonal ambiguity
of the problem; in short, there is no guarantee that $\hat{V}_1^i$ will be
sufficiently ``aligned'' with each other for their average to be close to $V_1$.
One must first resolve the orthogonal ambiguity in local solutions to
meaningfully aggregate them.

Garber et al.~\cite{GSS17} solve this problem when $r = 1$ in the setting of distributed PCA,
under fairly minimal assumptions. Specifically, they show that averaging a certain
combination of the local eigenvector estimates recovers a vector $\bar{v}_1$
that satisfies (with high probability)
\[
    \newstuff{\dist_2(\bar{v}_1, v_1) = \tilde{\cO}\left(\sqrt{\frac{1}{\delta^2 m n}} +
    \frac{1}{\delta^2 n}\right)},
\]
where $v_1$ is the leading eigenvector of $X$ and $\delta := \lambda_1(X) -
\lambda_2(X)$.
Letting $\hat{v}_1^{(i)}$ denote the estimate of $v_1$ produced by the $i\tsup{th}$
machine, the trick from Garber et al.~\cite{GSS17} is to pick a ``reference'' vector, e.g., $\hat{v}^1_1$,
and ``align'' all other local estimates with it to resolve the aforementioned
ambiguity --- which reduces to a sign ambiguity as $r = 1$.
More specifically, up to normalization, they compute $\bar{v}_1$ by the
``sign-fixed'' average:
\begin{equation}
	\bar{v}_1 := \frac{1}{m} \sum_{i = 1}^m
    \sign\bigg(\ip{\hat{v}_1^{(i)}, \hat{v}_1^{(1)}}\bigg) \hat{v}^{(i)}_1,
    \quad \hat{v}_1^{(i)} := \argmax_{v \in \Sbb^{d - 1}} v^{\T} \est^i v.
	\label{eq:sign-fixing}
\end{equation}
This is the same as \eqref{eq:basic_agg} for dimension $r = 1$, with
the additional sign-fixing in front of the estimate $\hat{v}^{(i)}_1$.
Conceptually, if we omitted the sign-fixing step, we would be averaging
eigenvectors that are ``spread'' around two directions: $\set{\pm v_1}$. If
each local estimate has the same probability of pointing to $v_1$ and $-v_1$,
the average will \newstuff{be no better at estimating $v_1$ than any local solution},
with error at least on the order of \newstuff{$\sqrt{\sfrac{1}{n}}$}~\cite[Theorem 3]{GSS17}.

\subsection{Procrustes fixing}
We propose a high-order analogue of the sign-fixing average over local
estimates. When $r > 1$, the canonical way to align two matrices
$\hat{V}, V \in \Obb_{n, r}$  is via the solution of the so-called
\textit{orthogonal Procrustes problem}~\cite[Chapter 6.4]{GVL13}:
\begin{equation}
	\argmin_{Z \in \Obb_r} \norm{\hat{V} - V Z}_F.
	\label{eq:procrustes}
\end{equation}
The alignment problem in~\eqref{eq:procrustes} admits a closed form solution~\cite{Higham1988}:
let $P \Sigma Q^{\T}$ be the SVD of \newstuff{$V^{\T} \hat{V}$}; then $Z := PQ^{\T}$
is the solution to~\eqref{eq:procrustes}.
Given the local estimates $\hat{V}_1^{(1)}, \dots, \hat{V}_1^{(m)}$ of the
principal eigenspace, we may choose one  of them to become the ``reference''
solution (e.g. $\hat{V}_1^{(1)}$) and return
$\tilde{V}$ defined below as our estimator:
\begin{equation}
	\tilde{V}, \tilde{R} := \texttt{qr}(\bar{V}), \quad
	\bar{V} := \frac{1}{m} \sum_{i=1}^m \hat{V}_1^{(i)} Z_i,
    \quad Z_i := \argmin_{Z \in \Obb_r} \norm{\hat{V}_1^{(i)} Z - \hat{V}_1^{(1)}}_F
	\label{eq:procrustes-fixing}
\end{equation}
In particular, note that~\eqref{eq:procrustes-fixing} recovers the sign-fixed
average of~\cref{eq:sign-fixing} when $r = 1$, since it follows that
\(
    \argmin_{s \in \set{\pm 1}} \norm{\hat{v}_1^{(i)} - s \hat{v}_1^1}_2 =
    \sign\left(\ip{\hat{v}_1^{(i)},\hat{v}_1^{(1)}}\right).
\)
Our algorithm is more formally described in \cref{alg:procrustes-fixing}.
By default, \cref{alg:procrustes-fixing} uses the first local
solution as a reference for~\eqref{eq:procrustes-fixing}; \newstuff{however,
since the order is arbitrary, our results are valid for any local solution used
as reference}.
In the experimental section, we demonstrate that iteratively refining the
reference solution $\hat{V}$ can further reduce the empirical error of the resulting estimator.

\begin{algorithm}[t]
	\caption{Distributed eigenspace estimation with Procrustes fixing}
	\begin{algorithmic}
        \State \textbf{Input}: local principal subspaces $\set{\hat{V}_1^{(i)} \mid
        i \in [m]}$, reference solution $\hat{V}$ (default: $\hat{V}_1^{(1)}$)
        \For{$i = 1, \cdots, m$}
            \State $\tilde{V}^{(i)} := \hat{V}^{(i)}_1 Z_i,
            \quad Z_i := \argmin_{Z \in \Obb_r} \norm{\hat{V}^{(i)}_1 Z -
            \hat{V}}_F$
        \EndFor
		\State Form $\bar{V} := \frac{1}{m} \sum_{i = 1}^m \tilde{V}^{(i)}$
        \State \Return $\tilde{V}$ from $\tilde{V}, \tilde{R} =
		\texttt{qr}(\bar{V})$.
	\end{algorithmic}
	\label{alg:procrustes-fixing}
\end{algorithm}

\newstuff{
\begin{remark}[Runtime comparison with~\cite{FWWZ19}]
    The algorithm proposed in~\cite{FWWZ19} for the distributed PCA problem
    works by spectral projector averaging. After local solutions have been
    transmitted to the central node, the algorithm therein requires
    computing the top $r$ eigenvectors of $\frac{1}{m} \sum_{i=1}^m
    \hat{V}_1^{(i)} (\hat{V}_1^{(i)})^{\T}$. Even forming this average would require
    roughly $\cO(m d^2 r)$ operations, which can be prohibitive; therefore, one
    would instead opt for a standard iterative algorithm such as orthogonal iteration.
    Each iteration would go through the following steps:
    \begin{enumerate}
        \item Compute $X \mapsto \hat{V}_1^{(i)}
            (\hat{V}_1^{(i)})^{\T} X$, for $X \in \Rbb^{d \times r}$
            and all $i \in [m]$. This step would require $\cO(mdr^2)$ operations,
            yielding a set of matrices of size $d \times r$ as intermediate results.
        \item Average the $m$ intermediate results, for a total of $\cO(mdr)$ operations.
        \item Orthogonalize the iterate (possibly every few iterations instead of every
            single iteration), incurring a cost $\cO(dr^2)$ operations (e.g. by using
            the QR factorization).
    \end{enumerate}
    \noindent Therefore, the cost of a \textbf{single step} of the iterative algorithm
    of choice would be $\cO(mr^2 d)$ operations in the central node, when using the
    method of~\cite{FWWZ19}.

    \noindent In contrast, and \textit{assuming no further parallelization}, our algorithm only needs to
    solve $m - 1$ Procrustes problems, which are equivalent to $m - 1$ SVDs of
    the matrices $(\hat{V}_1^{(1)})^{\T} \hat{V}_1^{(i)} \in \Rbb^{r \times r}$.
    Forming each of these matrices takes time $\cO(r^2 d)$, and computing the
    SVD would require time $\cO(r^3)$ (using e.g., the Golub-Kahan bidiagonalization
    method). The total cost therefore scales as $\cO(mr^2d)$.
\end{remark}
}

\newstuff{
\begin{remark}[Further parallelization]
    Note that~\cref{alg:procrustes-fixing} can be further parallelized at the
    expense of only a couple of additional communication rounds. In particular,
    the central node can first broadcast the reference solution $\hat{V}_1^{(1)}$
    to the rest of the $m - 1$ nodes, each of which solves the Procrustes problem
    locally and only transmits back the ``aligned'' solution $\tilde{V}^{(i)}$.
    The observed cost given this additional parallelization step is equal to
    \(
        \cO(T_{\mathsf{comm}} + r^2 d),
    \)
    where $T_{\mathsf{comm}}$ is the time required for a single round of communication.
\end{remark}}

Ideally, we would like to show that the performance of~\cref{alg:procrustes-fixing}
matches that of the centralized version. A na{\"i}ve attempt to adapt the
proof of~\cite{GSS17} poses a set of challenges: on one hand, the
case $r = 1$ relies on a Taylor-like expansion of each local estimate,
which is not transferrable to $r > 1$ in the absence of stronger assumptions
on the spectrum of the matrices involved. On the other hand, some of the arguments
in~\cite{GSS17} that deal with the ambiguity of the local estimates are no longer
longer applicable because of the presence of an arbitrary orthogonal transform, instead of
just a sign ambiguity.

Instead, we use recent results from numerical
analysis and matrix perturbation theory. First, we show that if the local
estimates $\hat{V}^{(i)}_1$ were \textit{already aligned with} $V_1$, in the
sense that
\begin{equation}
    \min_{Z \in \Obb_r} \norm{V_1 Z - \hat{V}^{(i)}_1}_F =
    \norm{V_1 - \hat{V}^{(i)}_1}_F,
\end{equation}
then averaging yields an estimate whose error is the sum of a term whose
magnitude is \textit{quadratic} in the local errors $\set{\norm{\est^i - X}_2 \mid
i \in [m]}$\footnote{These errors are typically $o(1)$.} plus
another term that depends on the error of the empirical mean approximation to $X$, i.e.,
$m^{-1}\sum_{i} \left(\est^i - X\right)$. To show this, we express $\hat{V}^{(i)}_1$
using $V_1$ as a local basis, a method that has been used to
analyze invariant subspace perturbations~\cite{KK14,DS19}.
Indeed, in the setting of distributed PCA, this yields an estimator whose error
matches that of the centralized estimator.

Recall that if the columns of $V_1$ form a basis for the leading invariant
subspace of $X$, so do those of $V_1 U$, where $U \in \Obb_r$ is any orthogonal
matrix. Thus we have a degree of freedom in choosing an appropriate ``version''
of $V_1$ to work with; without loss of generality, we choose $V_1$ so that
it minimizes the Procrustes distance to the first local solution $\hat{V}_1^{(1)}$;
in other words,
\begin{equation}
    \argmin_{U \in \Obb_r} \norm{\hat{V}_1^{(1)} U - V_1}_F = I_r.
    \label{eq:V1-aligned}
\end{equation}
Once the matrix $V_1$ is fixed,~\cref{alg:procrustes-fixing} aligns all other
local estimates with a solution ``sufficiently close'' to $V_1$. Indeed,
Stewart~\cite{Stewart12} showed that aligning $\hat{V}^{(i)}_1$ with
$\hat{V}^{(1)}_1$ is the same as aligning $\hat{V}^{(i)}_1$
with $V_1$, up to a \textit{quadratic} error term.

Before we state our generic result, we formalize our assumptions below:
\begin{assumption}
    \label{asm:deterministic}
    There is a collection of matrices $\set{\est^i \mid i \in [m]}$ as well as
    a reference matrix $X$ (with leading invariant subspace $V_1$) that satisfy
    the following:
    \begin{itemize}
        \item $\lambda_r(X) - \lambda_{r+1}(X) \geq \delta$, for some $\delta > 0$.
        \item the error matrices $E^i := \est^i - X$ satisfy
            \newstuff{$\norm{E^i}_2 < \frac{\delta}{8}$}, for all $i \in [m]$.
    \end{itemize}
\end{assumption}
In particular, we will see later that \cref{asm:deterministic} is satisfied with
high probability in the setting of distributed PCA with i.i.d.\ samples. We can now
state our deterministic result:
\begin{theorem}
    \label{theorem:procustes-fixing-works}
    Let $X \in \Rbb^{d \times d}$ have spectral decomposition~\eqref{eq:X-evdecomp}
    and let $\est^i, \; i \in [m]$ denote the local samples of $X$ with leading
    invariant subspaces $\hat{V}^{(i)}_1 \in \Obb_{d, r}$. Let~\cref{asm:deterministic}
    hold; then, if $\tilde{V}$ is the output of~\cref{alg:procrustes-fixing}, it
    satisfies the following error bound:
    \begin{equation}
        \dist_{2}(\tilde{V}, V_1) \lesssim
            \frac{1}{\delta^2} \max_{i \in [m]} \norm{\est^i - X}_2^2
            + \frac{1}{\delta} \bignorm{\frac{1}{m} \sum_{i=1}^m \est^i - X}_2.
        \label{eq:procrustes-fixing-error}
    \end{equation}
\end{theorem}
\newstuff{
\begin{remark}
    The error expression in~\eqref{eq:procrustes-fixing-error} naturally decomposes
    into two terms. The second term is the result of the Davis--Kahan theorem applied
    to quantify the distance of the top eigenspace of the empirical average
    $\frac{1}{m} \sum_{i} \est^i$ from the true eigenspace, and is precisely the error
    of a centralized algorithm that approximates $V_1$ by the top eigenspace of the
    empirical average of the local matrices.
    The first term in~\eqref{eq:procrustes-fixing-error} is an order of magnitude smaller than
    the error of approximating the leading eigenspace using just the local solution
    $\hat{V}^{(i)}_1$, as long as the individual errors are $o(1)$. In this case, the
    contribution of the first term in~\eqref{eq:procrustes-fixing-error} is negligible,
    and the total error is comparable to that of a centralized estimator.
\end{remark}}
In the next section, we give a more detailed outline of the proof of
\cref{theorem:procustes-fixing-works} that highlights the individual technical
components and how they fit together. The proofs of the individual components
are deferred to the supplementary material.

\subsection{Proof outline}
\label{sec:proof-outline}
The proof of~\cref{theorem:procustes-fixing-works} first analyzes the performance
of an idealized (but fictitious) version of~\cref{alg:procrustes-fixing} that uses
an ``ideal'' reference solution. \newstuff{In particular, we first ``fix'' a matrix $V_1$ with
orthogonal columns spanning the leading eigenspace of $X$ and assume (without loss
of generality, due to symmetry) that $V_1$ is already aligned with $\hat{V}_1^{(1)}$,
in the sense that it satisfies~\eqref{eq:V1-aligned}.
Then, we introduce a collection of fictitious iterates $\mtrv_1^{(i)}$ -- not to be
confused with the (unaligned) local solutions $\hat{V}_1^{(i)}$ -- which represent
precisely what the alignment algorithm would output if $V_1$ was used as a
reference solution. Then, we use the path-independence result of~\cite{Stewart12} to relate the
Procrustes-fixing estimator to this idealized version.}

For brevity, we use the following notation: if $V = \bmx{V_1 & V_2}$ is the matrix
of eigenvectors of $X$, with $V_1$ containing the principal eigenvectors, we write
for an arbitrary matrix $Z$:
\begin{equation}
    Z_{ij} := V_i^{\T} Z V_j, \; \; i, j \in \set{1, 2}.
\end{equation}
The first ingredient in our proof is a local expansion Lemma, motivated by the
arguments in~\cite{DS19}.
\begin{lemma}
    \label{lemma:local-expansion-1}
    Let~\cref{asm:deterministic} hold and choose $\mtrv_1^{(i)}$ to be \newstuff{
    the matrix whose columns are a basis of the leading invariant subspace of
    $\est^i$, and furthermore is maximally aligned with $V_1$, in the sense that}
    \begin{equation}
        \argmin_{U \in \Obb_r} \norm{\mtrv_1^{(i)} U - V_1}_F = I_r,
        \quad \forall\, i \in [m].
        \label{eq:nearest-matrix}
    \end{equation}
    Let $\mtrz^{(i)}$ be the root of the equation:\footnote{See~\cite[Section 3.2]{DS19} for motivation of this quantity.}
    \begin{equation}
        0 = -\est^i_{21} + \mtrz^{(i)} \est_{11}^i
        - \est_{22}^i \mtrz^{(i)} + \mtrz^{(i)} \est_{12}^i \mtrz^{(i)}.
        \label{eq:Z-root}
    \end{equation}
    and define $\mtry^{(i)} := V_2 \mtrz^{(i)}$ for $i \in [m]$.
    Then the following holds:
    \begin{equation}
        \norm{\mtrv_1^{(i)} - V_1 - \mtry^{(i)}}_2
        \lesssim \frac{\norm{\est^i - X}_2^2}{\delta^2}, \quad \forall\, i \in [m].
        \label{eq:local-expansion}
    \end{equation}
\end{lemma}
\begin{proof}
    See Section~\ref{sec:local-expansion-1-proof}.
\end{proof}
The expansion in~\cref{eq:local-expansion} is nearly sufficient for our
purposes; in particular, if we knew that $V_2 Z^{(i)}$ has small enough
spectral norm, summing over $i$ and applying the triangle
inequality would be sufficient. Even though this is not the case here,
we can still show that the average $m^{-1} \sum_i V_2 Z^{(i)}$ depends on the
approximation error of the empirical average $m^{-1} \sum_i \est^i - X$, as well
as the squares of the local approximation errors $\est^i - X$, which can
be studied and bounded on a per-application basis.
\begin{lemma}
    \label{lemma:local-expansion-sylvester}
    In the same setting as~\cref{lemma:local-expansion-1},
    the set of matrices $\mtry^{(i)}$ for $i \in [m]$ satisfy
    \begin{equation}
        \norm{\frac{1}{m} \sum_{i=1}^m \mtry^{(i)}}_2
        \lesssim \frac{1}{\delta^2 m} \sum_{i=1}^m \norm{\est^i - X}_2^2
        + \frac{1}{\delta} \norm{\frac{1}{m} \sum_{i=1}^m \est^i - X}_2
        \label{eq:Yhat-concentration}
    \end{equation}
\end{lemma}
\begin{proof}
    See~\Cref{sec:local-expansion-sylvester-proof}.
\end{proof}
With~\Cref{lemma:local-expansion-1,lemma:local-expansion-sylvester} at hand, we
can finally give an error bound for the average of the fictitious estimates
$\mtrv^{(i)}_1$.

\begin{proposition}
    \label{proposition:average-subspace}
    Let~\cref{asm:deterministic} hold and $\mtrv_1^{(i)}$ be
    defined as in~\cref{lemma:local-expansion-1}.
    Then the following holds:
    \begin{equation}
        \norm{\frac{1}{m} \sum_{i = 1}^m \mtrv_1^{(i)} - V_1}_2
        \lesssim \frac{1}{\delta^2 m}
        \sum_{i = 1}^m \norm{\est^i - X}_2^2 + \frac{1}{\delta}
        \norm{\frac{1}{m} \sum_{i = 1}^m \est^i - X}_2.
        \label{eq:average-subspace}
    \end{equation}
\end{proposition}
\begin{proof}
    Applying the triangle inequality and the above results,
    \begin{align}
        \label{eq:rev-triangle-expansion-2}
        \norm{\frac{1}{m} \sum_{i=1}^m \mtrv_1^{(i)} - V_1}_2
        &= \norm{ \frac{1}{m} \sum_{i=1}^m \mtrv_1^{(i)} - V_1 - \mtry^{(i)} + \mtry^{(i)} }_2 \\
        & \leq \frac{1}{m}\norm{ \sum_{i=1}^m \mtrv_1^{(i)} - V_1 - \mtry^{(i)} }_2  + \norm{ \frac{1}{m}\sum_{i=1}^{m}\mtry^{(i)} }_2 \\
        & \overset{\mathclap{(\text{\cref{lemma:local-expansion-1}})}}{\lesssim}
        \quad \frac{1}{\delta^2 m} \sum_{i = 1}^m \norm{\est^i - X}_2^2
        + \norm{ \frac{1}{m} \sum_{i=1}^m \mtry^{(i)} }_2 \\
        & \overset{\mathclap{(\text{\cref{lemma:local-expansion-sylvester}})}}{\lesssim}
        \quad
        \frac{1}{\delta^2 m} \sum_{ i = 1 }^m \norm{\est^i - X}_2^2
        + \frac{1}{\delta}
        \norm{\frac{1}{m} \sum_{i = 1}^m \est^i - X}_2.
    \end{align}
\end{proof}
This result holds for the fictitious estimates $\mtrv_1^{(i)}$;
however, it is not clear that the Procrustes-aligned estimates, $\tilde{V}^{(i)}$,
of~\cref{alg:procrustes-fixing} also satisfy such a property.
To show this, we leverage a recent result by Stewart~\cite{Stewart12} below;
informally, the result says that aligning with an ``accurate enough''
reference $\hat{V}_1^{(1)}$ is equivalent to directly aligning with $V_1$, up
to quadratic error in $\est^i - X$.
\begin{lemma}
    \label{lemma:path-independence}
    Let $\mtrv_1^{(i)}$ be defined as in
    \cref{proposition:average-subspace} and $\tilde{V}^{(i)}$ as
    in~\cref{alg:procrustes-fixing}. Then
    \begin{equation}
        \tilde{V}^{(i)} = \mtrv_1^{(i)} + T^{(i)}, \quad
        \norm{T^{(i)}}_2 \lesssim \max\set{
        \norm{\est^1 - X}_2^2, \norm{\est^i - \est^1}_2^2,
        \norm{\est^i - X}_2^2}.
        \label{eq:path-independence}
    \end{equation}
\end{lemma}
\begin{proof}
    See~\Cref{sec:path-independence-proof}.
\end{proof}

With Lemma~\ref{lemma:path-independence} at hand, we are now ready to
show that the Procrustes-fixing estimates produce a good approximation to the
leading invariant subspace $V_1$. Our results here do not depend
on a particular application setting such as distributed PCA; we are only using
matrix compuations and \cref{asm:deterministic}.
\begin{theorem}[Procrustes fixing]
    \label{theorem:procrustes-estimates}
    Let \cref{asm:deterministic} hold.
    The estimates $\tilde{V}^{(i)}$ from \cref{alg:procrustes-fixing} satisfy
    \begin{equation}
        \bignorm{\frac{1}{m} \sum_{i=1}^m \tilde{V}^{(i)} - V_1}_2
        \lesssim
        \frac{1}{\delta^2} \max_{ i \in [m] } \norm{\est^i - X}_2^2
                + \frac{1}{\delta}
                \bignorm{\frac{1}{m} \sum_{i = 1}^m \est^i - X}_2.
    \end{equation}
    Consequently, the output of Algorithm~\ref{alg:procrustes-fixing} satisfies
    \begin{equation}
        \dist_{2}(\tilde{V}, V_1) \lesssim
        \frac{1}{\delta^2} \max_{ i \in [m] } \norm{\est^i - X}_2^2
                + \frac{1}{\delta}
                \bignorm{\frac{1}{m} \sum_{i = 1}^m \est^i - X}_2.
    \end{equation}
\end{theorem}
\begin{proof}
    Under~\cref{asm:deterministic}, applying~\cref{lemma:path-independence} lets
    us rewrite $\tilde{V}^{(i)} = \mtrv^{(i)} + T^{(i)}$ where $T^{(i)}$ satisfies
    \begin{equation}
        \norm{T^{(i)}}_2 \lesssim \max\set{\norm{\est^i - X}_2^2,
            \norm{\est^i - \est^1}_2^2,
            \norm{\est^1 - X}_2^2}, \quad
            \forall i \in [m].
        \label{eq:Ti-bound}
    \end{equation}
    where we can further upper bound (via Young's inequality $(a + b)^2
    \leq 2(a^2 + b^2)$):
    \begin{align}
        \norm{\est^i - \est^1}^2_2 &= \norm{\est^i - X + X - \est^1}^2_2
        \lesssim \max\set{\norm{\est^i - X}_2^2, \norm{\est^1 - X}_2^2},
        \label{eq:Ti-bound-rewrite} \\
        \Rightarrow
        \norm{T^{(i)}}_2 &\lesssim \max_{i \in [m]} \norm{\est^i - X}_2^2.
        \label{eq:Ti-bound-complete}
    \end{align}
    By rewriting the desired statement and applying the triangle
    inequality, we have
    \begin{align}
        \bignorm{\frac{1}{m} \sum_{i=1}^m \tilde{V}^{(i)} - V_1}_2
            &\leq
        \bignorm{\frac{1}{m} \sum_{i=1}^m \mtrv^{(i)}_1 - V_1}_2
        + \frac{1}{m} \sum_{i=1}^m \norm{T^{(i)}}_2 \notag \\
            &\overset{\mathclap{(\text{Prop.~\ref{proposition:average-subspace}})}}{\lesssim}
            \quad
            \frac{1}{\delta^2 m} \sum_{i = 1}^m \norm{\est^i - X}_2^2
           + \frac{1}{\delta} \bignorm{\frac{1}{m} \sum_{i=1}^m \est^i - X}_2
        + \frac{1}{m} \sum_{i=1}^m \norm{T^{(i)}}_2 \notag \\
            &\overset{\mathclap{\eqref{eq:Ti-bound-complete}}}{\leq}
            \quad
        \frac{1}{\delta^2 m} \sum_{i = 1}^m \norm{\est^i - X}_2^2
           + \frac{1}{\delta} \bignorm{\frac{1}{m} \sum_{i=1}^m \est^i - X}_2 +
           \max_{i \in [m]} \norm{\est^i - X}_2^2 \notag \\
            &\lesssim \quad \frac{1}{\delta^2} \max_{i \in [m]} \norm{\est^i - X}_2^2
            + \frac{1}{\delta} \bignorm{\frac{1}{m} \sum_{i=1}^m \est^i - X}_2,
            \notag
    \end{align}
    which completes the proof.
\end{proof}

\subsection{Consequences for distributed PCA} \label{sec:distributed_PCA}
With~\cref{theorem:procrustes-estimates} at hand, we can show that applying
\cref{alg:procrustes-fixing} to the distributed PCA problem can yield estimates
whose error rate matches that of a centralized estimator. The first setting we
consider is similar to that of~\cite{GSS17}, summarized below.

\begin{assumption}
    \label{asm:problem-setting}
    Consider a zero-mean distribution $\cD$ supported on $\Rbb^d$, with covariance
    matrix $X = \expec[x \sim \cD]{xx^\T}$.
    We assume the following:
    \begin{enumerate}
        \item \label{item:asm-1}
            Each of the $m$ available machines draws $n$ i.i.d.\ samples from
            $\cD$, denoted $x_j^{(i)}$ for $j \in [n]$, $i \in [m]$.
        \item \label{item:asm-2}
            The population covariance matrix $X$ satisfies an eigengap
            condition:
            \begin{equation}
                \lambda_r(X) - \lambda_{r+1}(X) \geq \delta > 0,
                \label{eq:eiggap}
            \end{equation}
            where $\delta$ is some fixed scalar and $r$ is the rank
            of the target subspace.
        \item \label{item:asm-3}
            Any $x \sim \cD$ satisfies $\norm{x}_2 \leq \sqrt{b}$ almost surely.
    \end{enumerate}
\end{assumption}
Note that part~(\ref{item:asm-3}) of~\Cref{asm:problem-setting} is just for
simplicity, and adopted in~\cite{GSS17} as well. Later, we will show that
covariance matrices with small intrinsic dimension enable improved statistical
rates.

We first argue that \cref{asm:deterministic} is satisfied with high probability in this setting.
To do so, we first define the events
\begin{align}
    \cE &:= \cE_1 \cap \cE_2 \\
    \cE_1 &:= \set{\bigg\| \frac{1}{m} \sum_{i=1}^m \est^i - X \bigg\|_2
    \leq 2 \cdot \sqrt{\frac{\newstuff{b^2} \log(2 d / p)}{mn}}} \\
        \cE_2 &:= \set{\max_{i \in [m]} \norm{\est^i - X}_2 \leq \min\set{\frac{\delta}{8},
        2 \cdot \sqrt{\frac{\newstuff{b^2} \log (2d m / p)}{n}}}},
\end{align}
where $p$ is a parameter controlling the probability of failure and $\est^i :=
\frac{1}{n} \sum_{j=1}^n x_j^{(i)} (x_j^{(i)})^{\T}$ denote the local empirical
covariance matrices with leading $r$-dimensional invariant subspaces $\hat{V}^{(i)}_1$.
By~\cref{lemma:high-prob-events}, we have that
\begin{equation}
    \prob{\cE} \geq 1 - 2p - 2dm\expfun{-\frac{n\delta^2}{4 b^2}}.
    \label{eq:E-prob}
\end{equation}
Then we obtain the following Theorem:
\begin{theorem}
    \label{theorem:procrustes-pca}
    Under~\cref{asm:problem-setting}, the estimate $\tilde{V}$ returned
    by~\cref{alg:procrustes-fixing} satisfies
    \begin{equation}
        \dist_2(\tilde{V}, V_1) \lesssim
        \sqrt{\frac{b^2 \log(2 d / p)}{\delta^2 mn}} +
        \frac{b^2 \log(2 d m / p)}{\delta^2 n},
    \end{equation}
    with probability at least $1 - 2 p - 2dm \expfun{-\frac{n\delta^2}{4b^2}}$,
    \newstuff{whenever $n \gtrsim \log \frac{dm}{p}$}.
\end{theorem}


\Cref{theorem:procrustes-pca} shows that the error rate of distributed
PCA with~\cref{alg:procrustes-fixing} decays roughly as
$\sqrt{\sfrac{\newstuff{b^2} \log(c d/p)}{\delta^2 mn}}$, under the assumption
that $\norm{x_i}_2 \leq \sqrt{b}$ almost surely; however, typical modelling
choices, including the case where $x_i \sim \cN(0, \Sigma)$ which is explored
in the numerical experiments of~\Cref{sec:numerics-1}, do not guarantee such
boundedness. Although rates for covariance estimation of subgaussian vectors
are on the order of $\sqrt{\sfrac{d}{mn}}$~\cite[Chapter 6.3]{Wainwright19},
when the covariance matrix exhibits rapid spectral decay the statistical error
depends on its \textit{intrinsic dimension}, defined for a
PSD matrix $A$ as
\begin{equation}
    \textsf{intdim}(A) := \frac{\trace{A}}{\norm{A}_2} =
    1 + \sum_{i \geq 2} \frac{\lambda_i(A)}{\lambda_1(A)}.
    \label{eq:intdim}
\end{equation}
Notice that in general $1 \leq \textsf{intdim}(A) \leq \rank(A)$, and that
it is possible to have $\textsf{intdim}(A) \ll \rank(A)$. Using the intrinsic
dimension, we can refine our rates as follows:
\begin{theorem}
    \label{theorem:stablerank-rate}
    Modify~\cref{asm:problem-setting} so that~\cref{item:asm-3} (boundedness) is
    no longer required, and assume $\cD$ is a zero-mean subgaussian multivariate
    distribution with covariance matrix $X := \expec[x \sim \cD]{xx^{\T}}$.
    Let $r_{\star} := \mathsf{intdim}(X)$.
    Then, as long as $n \gtrsim \frac{r_{\star} + \log(m / p)}{\delta^2}$,
    the output of Algorithm~\ref{alg:procrustes-fixing} satisfies
    \begin{equation}
        \dist_2(\tilde{V}, V_1) \lesssim
        \frac{r_{\star} + \log(m / p)}{n} \cdot \left(\frac{\norm{X}_2}{\delta}\right)^2
        + \sqrt{\frac{r_{\star} + \log(c_1 n)}{mn}} \cdot \frac{\norm{X}_2}{\delta}
        \label{eq:better-concentration}
    \end{equation}
    with probability at least $1 - p - 2n^{-c_1}$, where $c_1$ is a
    dimension-independent constant.
\end{theorem}
\begin{proof}
    The proof uses the matrix deviation inequality~\cite{Vershynin18} to obtain
    tighter concentration for the empirical covariance matrices, but is otherwise
    identical to the proof of~\Cref{theorem:procustes-fixing-works}.
    See~\Cref{sec:stablerank-rate-proof} for details.
\end{proof}
\newstuff{The rate from~\cref{theorem:stablerank-rate} is similar to that of~\cite[Theorem 4]{FWWZ19},
absent an additional factor of $\sqrt{r}$ appearing in the latter. However, the
two bounds are not directly comparable; the error bound in~\cite{FWWZ19} is stated in
terms of the Frobenius distance of the spectral projectors $\dist_{F}(V, \hat{V}) :=
\norm{VV^{\T} - \hat{V}\hat{V}^{\T}}_F$.
It may also be possible to adapt the analysis of~\cite{FWWZ19} to remove the extra
$\sqrt{r}$ factor when the subspace distance is measured in the spectral norm.}
Table~\ref{tab:rate-comparison} gives a summary of statistical error rates in distributed PCA under different
settings.

\begin{table}[htbp]
    \centering
    \caption{Comparison of rates for distributed PCA. Here, $\kappa :=
        \frac{\norm{\Sigma}_2}{\delta}$, where $\delta$ is the population eigengap,
        $r$ is the dimension of the principal subspace, and $r_{\star} :=
    \mathsf{intdim}(\Sigma)$.}
    \begin{tabular}{c c r} \toprule
        \textbf{Setting} & \textbf{Rate} & \textbf{Reference} \\ \midrule \midrule
     \multirow{2}{*}{$\cD \subset \sqrt{b} \mathbf{B}^d$}
                         & \multirow{2}{*}{$\tilde{\cO}\left(\sqrt{\frac{\newstuff{b^2}}{\delta^2 mn}}
+ \frac{\newstuff{b^2}}{\delta^2 n}\right)$} & \cite{GSS17} $(r = 1)$ \\
                                     & & \cref{theorem:procrustes-pca} (general) \\ \midrule
     \multirow{2}{*}{$\cD$ subgaussian} &
    $\cO\left( \kappa \sqrt{\frac{r_{\star} + \log n}{mn}} + \kappa^2
\cdot \frac{r_{\star} + \log(m)}{n} \right)$ & \cref{theorem:stablerank-rate} \\
                                             & $\cO\left(
                                                     \sqrt{r} \kappa \sqrt{\frac{r_{\star}}{mn}} +
\sqrt{r} \kappa^2 \frac{r_{\star}}{n} \right) $ &
    \cite{FWWZ19}$^{\dagger}$  \\ \bottomrule
    \end{tabular}
    \label{tab:rate-comparison} \\

    \vspace{4pt}

    $\dagger$: Error bound given in terms of $\dist_F(V_1, \hat{V}) := \norm{V_1 V_1^{\T} - \hat{V}\hat{V}^{\T}}_F$.
\end{table}

\section{Numerical Experiments}
\label{sec:numerics}
In this section, we present numerical experiments on real and synthetic data.
The code to reproduce them, written in \texttt{Julia}, is available online.~\footnote{\url{https://gitlab.com/vchariso/distributed-eigenspace-estimation}}
In all experiments below, we use the label ``Central'' to refer to PCA using
the empirical covariance matrix of all $m \cdot n$ samples. In the synthetic
experiments, we use zero-mean multivariate Gaussians with covariance matrix
\begin{equation}
    \Sigma := U T U^{\T}, \; U \sim \text{Unif}(\Obb_d),
    \label{eq:covmat}
\end{equation}
where $T := \diag(\set{\tau_i}_{i \in [d]})$ is generated according to one
of the following two models (recall $r_{\star} := \mathsf{intdim}(\Sigma)$):
\begin{align}
    \tag{M1}
    \tau_i &:= \begin{cases}
        \lambda_h - \frac{(\lambda_h - \lambda_{\ell}) \cdot (i - 1)}{r - 1}, & i \leq r \\
        \left( \lambda_{\ell} - \delta \right) \cdot 0.9^{i - r - 1}, & i > r,
    \end{cases}, \quad \text{ or } \label{eq:covmat-construction} \\
    \tag{M2}
    \tau_i &:= \begin{cases}
        1, & i \leq r \\
        (1 - \delta) \cdot \alpha^{i - r}, & i > r
    \end{cases},
    \quad \text{where $\alpha$ solves} \quad
    \frac{(1 - \delta)}{1 - \alpha} = r_{\star} - r
    \label{eq:intdim-experiment}
\end{align}
In model~\eqref{eq:covmat-construction}, the $r$ principal eigenvalues are linearly
spaced in $[\lambda_{\ell}, \lambda_{h}]$. In contrast, in~\eqref{eq:intdim-experiment}
all principal eigenvalues are $1$, while the trailing eigenvalues decay according to
a specified rate $\alpha < 1$. Both constructions ensure that the eigengap is
exactly equal to $\delta$.

\subsection{Performance as a function of $m$ and $n$} \label{sec:numerics-1}
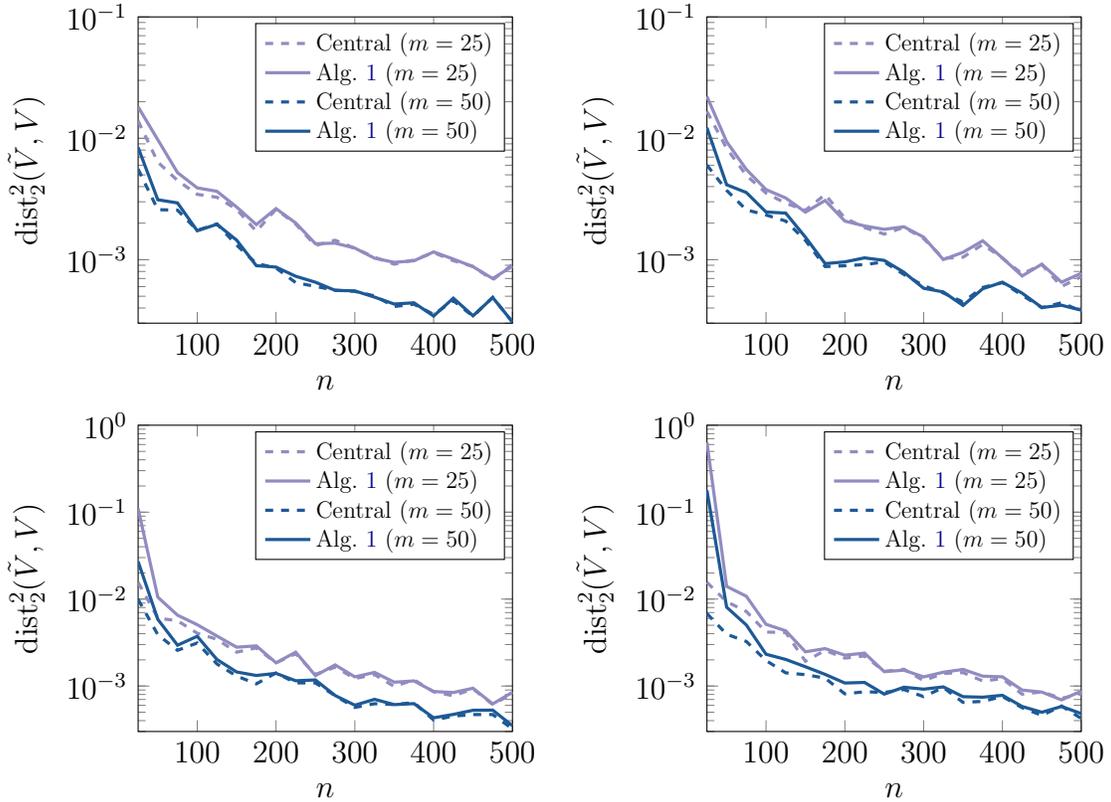
\begin{figure}[tb]
	\centering
	\begin{minipage}{0.45\textwidth}
	\begin{tikzpicture}
	\begin{axis}[xlabel=$n$,ylabel=${\dist_2^2(\tilde{V}, V)}$, ymode=log,
				 enlargelimits=false, width=0.9\linewidth,
                 legend cell align=left, ymax=0.1, ymin=3e-4,
				 legend style={nodes={scale=0.75, transform shape}}]
	\addplot[no markers, dashed, very thick, color=lblue]
	table[x=n,y=erm25, col sep=comma] {fig_data/error_normal-1_reps-5.csv};
	\addplot[no markers, very thick, color=lblue]
	table[x=n,y=fix25, col sep=comma]
	{fig_data/error_normal-1_reps-5.csv};
	\addplot[dashed, no markers, very thick, color=hblue]
	table[x=n,y=erm50, col sep=comma] {fig_data/error_normal-1_reps-5.csv};
	\addplot[no markers, very thick, color=hblue]
	table[x=n,y=fix50, col sep=comma] {fig_data/error_normal-1_reps-5.csv};
	\legend{Central ${(m = 25)}$, Alg.~\ref{alg:procrustes-fixing} ${(m = 25)}$,
		Central ${(m = 50)}$, Alg.~\ref{alg:procrustes-fixing} ${(m = 50)}$};
	\end{axis}
	\end{tikzpicture}
	\end{minipage}~
	\begin{minipage}{0.45\textwidth}
	\begin{tikzpicture}
	\begin{axis}[xlabel=$n$,ylabel=${\dist_2^2(\tilde{V}, V)}$, ymode=log,
				 enlargelimits=false, width=0.9\linewidth,
                 legend cell align=left, ymax=0.1, ymin=3e-4,
				 legend style={nodes={scale=0.75, transform shape}}]
	\addplot[dashed, no markers, very thick, color=lblue]
	table[x=n,y=erm25, col sep=comma] {fig_data/error_normal-4_reps-5.csv};
	\addplot[no markers, very thick, color=lblue]
	table[x=n,y=fix25, col sep=comma] {fig_data/error_normal-4_reps-5.csv};
	\addplot[dashed, no markers, very thick, color=hblue]
	table[x=n,y=erm50, col sep=comma] {fig_data/error_normal-4_reps-5.csv};
	\addplot[no markers, very thick, color=hblue]
	table[x=n,y=fix50, col sep=comma] {fig_data/error_normal-4_reps-5.csv};
	\legend{Central ${(m = 25)}$, Alg.~\ref{alg:procrustes-fixing} ${(m = 25)}$,
		Central ${(m = 50)}$, Alg.~\ref{alg:procrustes-fixing} ${(m = 50)}$};
	\end{axis}
	\end{tikzpicture}
	\end{minipage}\\
	\begin{minipage}{0.45\textwidth}
	\begin{tikzpicture}
	\begin{axis}[xlabel=$n$,ylabel=${\dist_2^2(\tilde{V}, V)}$, ymode=log,
	enlargelimits=false, width=0.9\linewidth, legend cell align=left,
	legend style={nodes={scale=0.75, transform shape}},
    ymax=1.0, ymin=3e-4]
	\addplot[dashed, no markers, very thick, color=lblue]
	table[x=n,y=erm25, col sep=comma] {fig_data/error_normal-8_reps-5.csv};
	\addplot[no markers, very thick, color=lblue]
	table[x=n,y=fix25, col sep=comma] {fig_data/error_normal-8_reps-5.csv};
	\addplot[dashed, no markers, very thick, color=hblue]
	table[x=n,y=erm50, col sep=comma] {fig_data/error_normal-8_reps-5.csv};
	\addplot[no markers, very thick, color=hblue]
	table[x=n,y=fix50, col sep=comma] {fig_data/error_normal-8_reps-5.csv};
	\legend{Central ${(m = 25)}$, Alg.~\ref{alg:procrustes-fixing} ${(m = 25)}$,
		Central ${(m = 50)}$, Alg.~\ref{alg:procrustes-fixing} ${(m = 50)}$};
	\end{axis}
	\end{tikzpicture}
\end{minipage}~
\begin{minipage}{0.45\textwidth}
	\begin{tikzpicture}
	\begin{axis}[xlabel=$n$,ylabel=${\dist_2^2(\tilde{V}, V)}$, ymode=log,
	enlargelimits=false, width=0.9\linewidth, legend cell align=left,
	legend style={nodes={scale=0.75, transform shape}},
    mark options={solid},
    ymax=1.0, ymin=3e-4]
	\addplot[dashed, no markers, very thick, color=lblue]
	table[x=n,y=erm25, col sep=comma] {fig_data/error_normal-16_reps-5.csv};
	\addplot[no markers, very thick, color=lblue]
	table[x=n,y=fix25, col sep=comma] {fig_data/error_normal-16_reps-5.csv};
	\addplot[dashed, no markers, very thick, color=hblue]
	table[x=n,y=erm50, col sep=comma] {fig_data/error_normal-16_reps-5.csv};
	\addplot[no markers, very thick, color=hblue]
	table[x=n,y=fix50, col sep=comma] {fig_data/error_normal-16_reps-5.csv};
	\legend{Central ${(m = 25)}$, Alg.~\ref{alg:procrustes-fixing} ${(m = 25)}$,
		Central ${(m = 50)}$, Alg.~\ref{alg:procrustes-fixing} ${(m = 50)}$};
	\end{axis}
	\end{tikzpicture}
\end{minipage}
\caption{Performance of centralized vs.\ distributed PCA with Procrustes-fixing
	for $m \in \set{25, 50}$ and $n \in \set{25, 50, \dots, 500}$, with $\delta
    = 0.2$. Target ranks (from left to right): $r \in \set{1, 4}$ (\textbf{top}),
    $r \in \set{8, 16}$ (\textbf{bottom}).}
\label{fig:procFixing}
\end{figure}

We generate a set of synthetic experiments following
the model~\eqref{eq:covmat-construction}, setting $\lambda_{\ell} = 0.5$,
$\lambda_{h} = 1$, and $\delta = 0.2$.
We then apply~\cref{alg:procrustes-fixing} for a variety of $(m, n)$ after
setting $d = 300$, which is the parameter used in~\cite{GSS17} (wherein the
case $r = 1$ is studied). Figure~\ref{fig:procFixing} depicts the performance of
Algorithm~\ref{alg:procrustes-fixing} for a number of different configurations.
For $r \in \set{1, 4, 8, 16}$, Algorithm~\ref{alg:procrustes-fixing} performs
essentially as well as a centralized PCA, which uses all $m \cdot n$ samples to form
the empirical covariance matrix. For $r = 1$, our error plots closely resemble
those from~\cite{GSS17}. In all configurations, ``na{\"i}ve'' averaging produces
an estimate $\bar{V}$ with error on the order of $\Omega(1)$, that does not always
decay as a function of $n$; for that reason, we omit its depiction.

\begin{figure}[tb]
	\centering
	\begin{minipage}{0.475\textwidth}
	\begin{tikzpicture}
	\begin{axis}[xlabel=$m$,ylabel=${\dist_2^2(\tilde{V}, V)}$, ymode=log,
				 enlargelimits=false, width=0.85\linewidth,
                 legend cell align=left, legend pos=north west,
                 mark options={solid},
                 ymax=5e-3, legend style={nodes={scale=0.75, transform shape}},
                 legend entries={
                     {$r = 2$}, {$r = 4$},
                     {Central}, {Alg.~\ref{alg:procrustes-fixing}},
                     {Alg.~\ref{alg:procrustes-fixing-iterative}}
                 }]
    \addlegendimage{solid, very thick, color=lblue};
    \addlegendimage{solid, very thick, color=hblue};
    \addlegendimage{mark=triangle, dashed, very thick, color=black};
    \addlegendimage{mark=square, very thick, color=black};
    \addlegendimage{mark=o, very thick, color=black};
    \addplot[mark=triangle, dashed, very thick, color=lblue]
        table[x=m,y=erm, col sep=comma] {fig_data/error_machines-2_reps-5.csv};
    \addplot[mark=square, very thick, color=lblue]
        table[x=m,y=fix, col sep=comma] {fig_data/error_machines-2_reps-5.csv};
    \addplot[mark=o, very thick, color=lblue]
        table[x=m,y=itr, col sep=comma] {fig_data/error_machines-2_reps-5.csv};
    \addplot[mark=triangle, dashed, very thick, color=hblue]
        table[x=m,y=erm, col sep=comma] {fig_data/error_machines-4_reps-5.csv};
    \addplot[mark=square, very thick, color=hblue]
        table[x=m,y=fix, col sep=comma] {fig_data/error_machines-4_reps-5.csv};
    \addplot[mark=o, very thick, color=hblue]
        table[x=m,y=itr, col sep=comma] {fig_data/error_machines-4_reps-5.csv};
	\end{axis}
	\end{tikzpicture}
	\end{minipage}\quad
	\begin{minipage}{0.475\textwidth}
	\begin{tikzpicture}
	\begin{axis}[xlabel=$m$,ylabel=${\dist_2^2(\tilde{V}, V)}$, ymode=log,
				 enlargelimits=false, width=0.85\linewidth,
                 legend cell align=left, mark options={solid},
                 legend pos=north west,
				 legend style={nodes={scale=0.75, transform shape}},
                 legend entries={
                     {$r = 8$}, {$r = 16$},
                     {Central}, {Alg.~\ref{alg:procrustes-fixing}},
                     {Alg.~\ref{alg:procrustes-fixing-iterative}}
                 }]
    \addlegendimage{solid, very thick, color=lblue};
    \addlegendimage{solid, very thick, color=hblue};
    \addlegendimage{mark=triangle, dashed, very thick, color=black};
    \addlegendimage{mark=square, very thick, color=black};
    \addlegendimage{mark=o, very thick, color=black};
    \addplot[mark=triangle, dashed, very thick, color=lblue]
        table[x=m,y=erm, col sep=comma] {fig_data/error_machines-8_reps-5.csv};
    \addplot[mark=square, very thick, color=lblue]
        table[x=m,y=fix, col sep=comma] {fig_data/error_machines-8_reps-5.csv};
    \addplot[mark=o, very thick, color=lblue]
        table[x=m,y=itr, col sep=comma] {fig_data/error_machines-8_reps-5.csv};
    \addplot[mark=triangle, dashed, very thick, color=hblue]
        table[x=m,y=erm, col sep=comma] {fig_data/error_machines-16_reps-5.csv};
    \addplot[mark=square, very thick, color=hblue]
        table[x=m,y=fix, col sep=comma] {fig_data/error_machines-16_reps-5.csv};
    \addplot[mark=o, very thick, color=hblue]
        table[x=m,y=itr, col sep=comma] {fig_data/error_machines-16_reps-5.csv};
	\addplot[mark=triangle, dashed, very thick, color=lblue]
        table[x=m,y=erm, col sep=comma] {fig_data/error_machines-8_reps-5.csv};
	\addplot[mark=square, very thick, color=lblue]
        table[x=m,y=fix, col sep=comma] {fig_data/error_machines-8_reps-5.csv};
    \addplot[mark=triangle, dashed, very thick, color=hblue]
        table[x=m,y=erm, col sep=comma] {fig_data/error_machines-16_reps-5.csv};
    \addplot[mark=square, very thick, color=hblue]
        table[x=m,y=fix, col sep=comma] {fig_data/error_machines-16_reps-5.csv};
	\end{axis}
	\end{tikzpicture}
	\end{minipage}
    \caption{Performance of centralized vs.\ distributed PCA with Procrustes-fixing
    for fixed $n \cdot m = 20000$ with $\delta = 0.2$ and varying $m$; here,
    Alg.~\ref{alg:procrustes-fixing-iterative} is invoked with $\texttt{n\_iter} = 2$.
    When the number of machines is too large, the accuracy of each individual solution
    degrades, leading to loss of accuracy in the averaged solution itself.}
\label{fig:procFixing-m}
\end{figure}
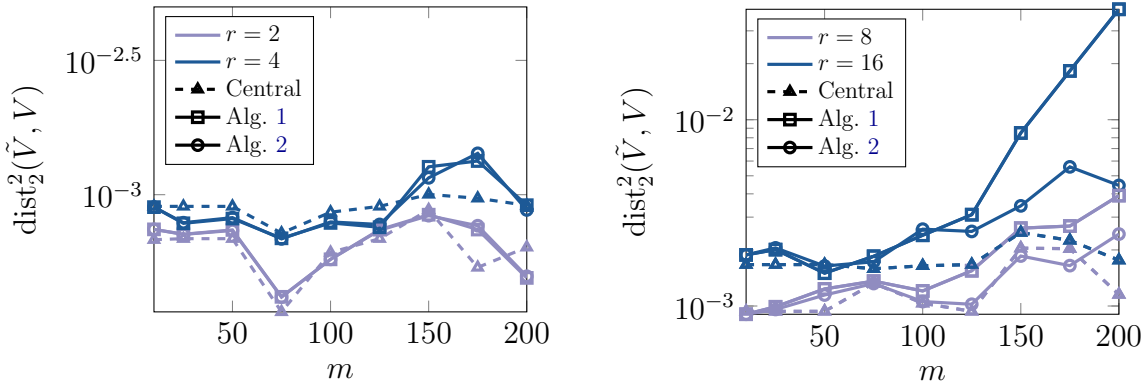

Crucially, our theory requires $n$ to be sufficiently large, since otherwise
certain conditions may not hold with high probability. For example, if $n$ is
too small, it is unlikely that $\norm{\est^i - X}_2 \leq \frac{\delta}{8}$, as
required. For that reason, we generate another synthetic experiment following
model~\eqref{eq:covmat-construction}. This time, we keep the total number of
samples $m \cdot n$ \textbf{fixed} and vary $m$. The results are shown in
\cref{fig:procFixing-m}. Unsurprisingly, larger values of $m$ lead to less
accurate local solutions and smaller overall accuracy of the
Procrustes-aligned solution. Note that even if most local solutions are
accurate enough, larger values of $m$ (or equivalently smaller values of $n$)
imply there is a higher probability that the
reference solution $\hat{V}_1^{(1)}$ is a poor approximation of $V_1$, which
inevitably leads to loss of accuracy during the alignment step.

\begin{algorithm}[tbp]
    \caption{Procrustes fixing with iterative refinement}
    \begin{algorithmic}
        \State \textbf{Input}: local principal subspaces $\set{\hat{V}_1^{(i)} \mid
        i \in [m]}$, number of refinement steps $\texttt{n\_iter}$
        \State Set $\tilde{V}^{(0)} := \hat{V}_1^{(1)}$
        \For{$k = 1, \dots, \texttt{n\_iter}$}
            \State compute $\tilde{V}^{(k)}$ using \cref{alg:procrustes-fixing}
                with reference solution $\tilde{V}^{(k - 1)}$
        \EndFor
        \State \Return $\tilde{V}^{(\texttt{n\_iter})}$.
    \end{algorithmic}
    \label{alg:procrustes-fixing-iterative}
\end{algorithm}

\subsection{Iterative refinement}
\label{sec:iterative-refinement}
The estimates generated by~\cref{alg:procrustes-fixing} \newstuff{were empirically
observed to be sensitive to the individual solution $\hat{V}_1^{(1)}$ chosen as a reference},
especially when the ratio $\sqrt{\sfrac{1}{n}}$ is non-negligible. To obtain a
more robust estimate, we propose a practical iterative refinement step
outlined in~\cref{alg:procrustes-fixing-iterative}. Instead of performing a single
round of Procrustes alignment, the algorithm goes through multiple iterations
where the output of the previous alignment round is used as a reference
solution for the current round. Intuitively, even if the initial reference
solution was somewhat inaccurate, the averaging step will likely smooth out
some of the error, leading to a more accurate reference solution.

\begin{figure}[!tbp]
    \centering
    \textbf{Comparison of Algorithms~\ref{alg:procrustes-fixing}
        and~\ref{alg:procrustes-fixing-iterative}}
    \vspace{10pt}
    \begin{minipage}{0.45\textwidth}
    \begin{tikzpicture}
        \begin{axis}[xlabel=$n$,ylabel=${\dist_2(\tilde{V}, V)}$,
                    enlargelimits=false, width=0.9\linewidth, legend cell align=left,
                    xmode=log, legend style={nodes={scale=0.85, transform shape}},
                    mark options={solid}, log basis x={2},
                    legend entries={{$r_{\star} = 4$}, {$r_{\star} = 8$}}]
            \addlegendimage{no markers, very thick, dashed};
            \addlegendimage{no markers, very thick, solid};
            \addplot[mark=triang;e, dashed, color=black, very thick]
                table[x=n, y=erm, col sep=comma] {fig_data/error_niter-2_srank-4-gap-0.10.csv};
            \addplot[mark=square, dashed, color=hblue, very thick]
                table[x=n, y=fix, col sep=comma] {fig_data/error_niter-2_srank-4-gap-0.10.csv};
            \addplot[mark=square, dashed, color=lblue, very thick]
                table[x=n, y=itr, col sep=comma] {fig_data/error_niter-2_srank-4-gap-0.10.csv};
            \addplot[mark=triangle, color=black, very thick]
                table[x=n, y=erm, col sep=comma] {fig_data/error_niter-2_srank-8-gap-0.10.csv};
            \addplot[mark=square, color=hblue, very thick]
                table[x=n, y=fix, col sep=comma] {fig_data/error_niter-2_srank-8-gap-0.10.csv};
            \addplot[mark=square, color=lblue, very thick]
                table[x=n, y=itr, col sep=comma] {fig_data/error_niter-2_srank-8-gap-0.10.csv};
        \end{axis}
    \end{tikzpicture}
    \end{minipage}~
    \begin{minipage}{0.45\textwidth}
    \begin{tikzpicture}
        \begin{axis}[xlabel=$n$,
                    enlargelimits=false, width=0.9\linewidth, legend cell align=left,
                    xmode=log, legend style={nodes={scale=0.85, transform shape}},
                    mark options={solid}, log basis x={2},
                    legend entries={{$r_{\star} = 16$}, {$r_{\star} = 32$}}]
            \addlegendimage{no markers, very thick, dashed};
            \addlegendimage{no markers, very thick, solid};
            \addplot[mark=triangle, dashed, color=black, very thick]
                table[x=n, y=erm, col sep=comma] {fig_data/error_niter-2_srank-16-gap-0.10.csv};
            \addplot[mark=square, dashed, color=hblue, very thick]
                table[x=n, y=fix, col sep=comma] {fig_data/error_niter-2_srank-16-gap-0.10.csv};
            \addplot[mark=square, dashed, color=lblue, very thick]
                table[x=n, y=itr, col sep=comma] {fig_data/error_niter-2_srank-16-gap-0.10.csv};
            \addplot[mark=triangle, color=black, very thick]
                table[x=n, y=erm, col sep=comma] {fig_data/error_niter-2_srank-32-gap-0.10.csv};
            \addplot[mark=square, color=hblue, very thick]
                table[x=n, y=fix, col sep=comma] {fig_data/error_niter-2_srank-32-gap-0.10.csv};
            \addplot[mark=square, color=lblue, very thick]
                table[x=n, y=itr, col sep=comma] {fig_data/error_niter-2_srank-32-gap-0.10.csv};
        \end{axis}
    \end{tikzpicture}
    \end{minipage}
    \begin{minipage}{0.45\textwidth}
    \begin{tikzpicture}
        \begin{axis}[xlabel=$n$,ylabel=${\dist_2(\tilde{V}, V)}$,
                    enlargelimits=false, width=0.9\linewidth, legend cell align=left,
                    xmode=log, legend style={nodes={scale=0.85, transform shape}},
                    mark options={solid}, log basis x={2},
                    legend entries={{$r_{\star} = 4$}, {$r_{\star} = 8$}}]
            \addlegendimage{no markers, very thick, dashed};
            \addlegendimage{no markers, very thick, solid};
            \addplot[mark=triangle, dashed, color=black, very thick]
                table[x=n, y=erm, col sep=comma] {fig_data/error_niter-5_srank-4-gap-0.10.csv};
            \addplot[mark=square, dashed, color=hblue, very thick]
                table[x=n, y=fix, col sep=comma] {fig_data/error_niter-5_srank-4-gap-0.10.csv};
            \addplot[mark=square, dashed, color=lblue, very thick]
                table[x=n, y=itr, col sep=comma] {fig_data/error_niter-5_srank-4-gap-0.10.csv};
            \addplot[mark=triangle, color=black, very thick]
                table[x=n, y=erm, col sep=comma] {fig_data/error_niter-5_srank-8-gap-0.10.csv};
            \addplot[mark=square, color=hblue, very thick]
                table[x=n, y=fix, col sep=comma] {fig_data/error_niter-5_srank-8-gap-0.10.csv};
            \addplot[mark=square, color=lblue, very thick]
                table[x=n, y=itr, col sep=comma] {fig_data/error_niter-5_srank-8-gap-0.10.csv};
        \end{axis}
    \end{tikzpicture}
    \end{minipage}~
    \begin{minipage}{0.45\textwidth}
    \begin{tikzpicture}
        \begin{axis}[xlabel=$n$,
                    enlargelimits=false, width=0.9\linewidth, legend cell align=left,
                    xmode=log, legend style={nodes={scale=0.85, transform shape}},
                    mark options={solid}, log basis x={2},
                    legend entries={{$r_{\star} = 16$}, {$r_{\star} = 32$}}]
            \addlegendimage{no markers, very thick, dashed};
            \addlegendimage{no markers, very thick, solid};
            \addplot[mark=triangle, dashed, color=black, very thick]
                table[x=n, y=erm, col sep=comma] {fig_data/error_niter-5_srank-16-gap-0.10.csv};
            \addplot[mark=square, dashed, color=hblue, very thick]
                table[x=n, y=fix, col sep=comma] {fig_data/error_niter-5_srank-16-gap-0.10.csv};
            \addplot[mark=square, dashed, color=lblue, very thick]
                table[x=n, y=itr, col sep=comma] {fig_data/error_niter-5_srank-16-gap-0.10.csv};
            \addplot[mark=triangle, color=black, very thick]
                table[x=n, y=erm, col sep=comma] {fig_data/error_niter-5_srank-32-gap-0.10.csv};
            \addplot[mark=square, color=hblue, very thick]
                table[x=n, y=fix, col sep=comma] {fig_data/error_niter-5_srank-32-gap-0.10.csv};
            \addplot[mark=square, color=lblue, very thick]
                table[x=n, y=itr, col sep=comma] {fig_data/error_niter-5_srank-32-gap-0.10.csv};
        \end{axis}
    \end{tikzpicture}
    \end{minipage}
    \begin{minipage}{0.45\textwidth}
    \begin{tikzpicture}
        \begin{axis}[xlabel=$n$,ylabel=${\dist_2(\tilde{V}, V)}$,
                    enlargelimits=false, width=0.9\linewidth, legend cell align=left,
                    xmode=log, legend style={nodes={scale=0.85, transform shape}},
                    mark options={solid}, log basis x={2},
                    legend entries={{$r_{\star} = 4$}, {$r_{\star} = 8$}}]
            \addlegendimage{no markers, very thick, dashed};
            \addlegendimage{no markers, very thick, solid};
            \addplot[mark=triangle, dashed, color=black, very thick]
                table[x=n, y=erm, col sep=comma] {fig_data/error_niter-15_srank-4-gap-0.10.csv};
            \addplot[mark=square, dashed, color=hblue, very thick]
                table[x=n, y=fix, col sep=comma] {fig_data/error_niter-15_srank-4-gap-0.10.csv};
            \addplot[mark=square, dashed, color=lblue, very thick]
                table[x=n, y=itr, col sep=comma] {fig_data/error_niter-15_srank-4-gap-0.10.csv};
            \addplot[mark=triangle, color=black, very thick]
                table[x=n, y=erm, col sep=comma] {fig_data/error_niter-15_srank-8-gap-0.10.csv};
            \addplot[mark=square, color=hblue, very thick]
                table[x=n, y=fix, col sep=comma] {fig_data/error_niter-15_srank-8-gap-0.10.csv};
            \addplot[mark=square, color=lblue, very thick]
                table[x=n, y=itr, col sep=comma] {fig_data/error_niter-15_srank-8-gap-0.10.csv};
        \end{axis}
    \end{tikzpicture}
    \end{minipage}~
    \begin{minipage}{0.45\textwidth}
    \begin{tikzpicture}
        \begin{axis}[xlabel=$n$,
                    enlargelimits=false, width=0.9\linewidth, legend cell align=left,
                    xmode=log, legend style={nodes={scale=0.85, transform shape}},
                    mark options={solid}, log basis x={2},
                    legend entries={{$r_{\star} = 16$}, {$r_{\star} = 32$}}]
            \addlegendimage{no markers, very thick, dashed};
            \addlegendimage{no markers, very thick, solid};
            \addplot[mark=triangle, dashed, color=black, very thick]
                table[x=n, y=erm, col sep=comma] {fig_data/error_niter-15_srank-16-gap-0.10.csv};
            \addplot[mark=square, dashed, color=hblue, very thick]
                table[x=n, y=fix, col sep=comma] {fig_data/error_niter-15_srank-16-gap-0.10.csv};
            \addplot[mark=square, dashed, color=lblue, very thick]
                table[x=n, y=itr, col sep=comma] {fig_data/error_niter-15_srank-16-gap-0.10.csv};
            \addplot[mark=triangle, color=black, very thick]
                table[x=n, y=erm, col sep=comma] {fig_data/error_niter-15_srank-32-gap-0.10.csv};
            \addplot[mark=square, color=hblue, very thick]
                table[x=n, y=fix, col sep=comma] {fig_data/error_niter-15_srank-32-gap-0.10.csv};
            \addplot[mark=square, color=lblue, very thick]
                table[x=n, y=itr, col sep=comma] {fig_data/error_niter-15_srank-32-gap-0.10.csv};
        \end{axis}
    \end{tikzpicture}
    \end{minipage}
    \begin{tikzpicture}
        \begin{axis}[%
            hide axis,
            xmin=10,
            xmax=50,
            ymin=0,
            ymax=0.4,
            legend style={draw=white!15!black,legend cell align=left,
                          nodes={scale=0.85, transform shape}},
            legend columns=-1,
            legend entries={
                {Alg.~\ref{alg:procrustes-fixing} \qquad},
                {Alg.~\ref{alg:procrustes-fixing-iterative} \qquad},
                {Central}}]
        \addlegendimage{mark=square, very thick, color=hblue};
        \addlegendimage{mark=square, solid, very thick, color=lblue};
        \addlegendimage{mark=triangle, solid, very thick, color=black};
        \end{axis}
    \end{tikzpicture}
    \caption{Empirical error of~\cref{alg:procrustes-fixing} vs.\
        \cref{alg:procrustes-fixing-iterative} for a problem with $d = 300, m = 50$
        and $\delta = 0.1$ for a variety of $n$ and $r_{\star} := \textsf{intdim}(X)$.
        $\texttt{n\_iter} = 2$ (\textbf{top}), $\texttt{n\_iter} = 5$ (\textbf{middle})
        and $\texttt{n\_iter} = 15$ (\textbf{bottom}). The problem instances
        generated across rows are identical.
        In the challenging regime where $n$ is small, iterative refinement can
        significantly reduce the estimation error. Smaller number of samples
        per machine $n$ always leads to more significant gaps between central
    and distributed solutions.}
    \label{fig:iterative-refinement}
\end{figure}
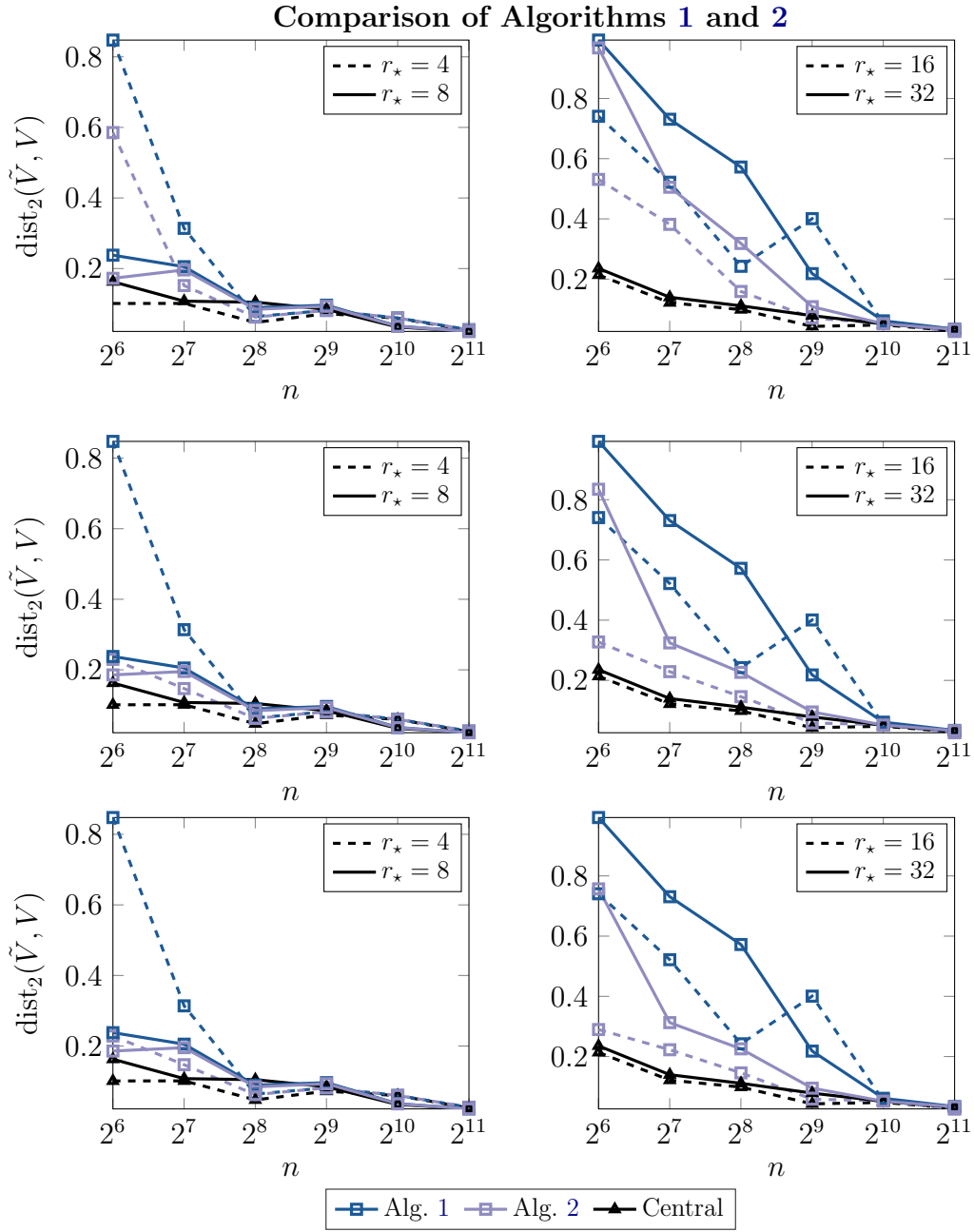

\newstuff{
To verify this, we perform two experiments. Initially, we compare the performance of Alg.~\ref{alg:procrustes-fixing}
and~\ref{alg:procrustes-fixing-iterative} (with $\texttt{n\_iter} = 2$) in the experiment
of~\cref{sec:numerics-1} where $m \cdot n$ is kept fixed and $m$ varies. Empirically,
we observe that the additional averaging steps make the most difference for small $n$,
where individual reference solutions are more likely to be innacurate.
Additionally, we compare~\Cref{alg:procrustes-fixing,alg:procrustes-fixing-iterative} using
synthetic instances generated by model~\eqref{eq:intdim-experiment} to examine the effect of
additional rounds of iterative refinement on the accuracy of solutions; the results can be found
in~\cref{fig:iterative-refinement}. Therein, we report the subspace distance from the ground truth
by varying $\texttt{n\_iter} \in \set{2, 5, 15}$.
For small $n$, iterative refinement can decrease the subspace distance significantly, at the cost of a
few extra rounds of computation. Moreover, just a few rounds of refinement
are sufficient; the difference between $5$ and $15$ refinement steps is
negligible in all configurations of $(n, r_{\star})$.}

\subsection{Performance as a function of intrinsic dimension}
\label{sec:intdim-experiment}
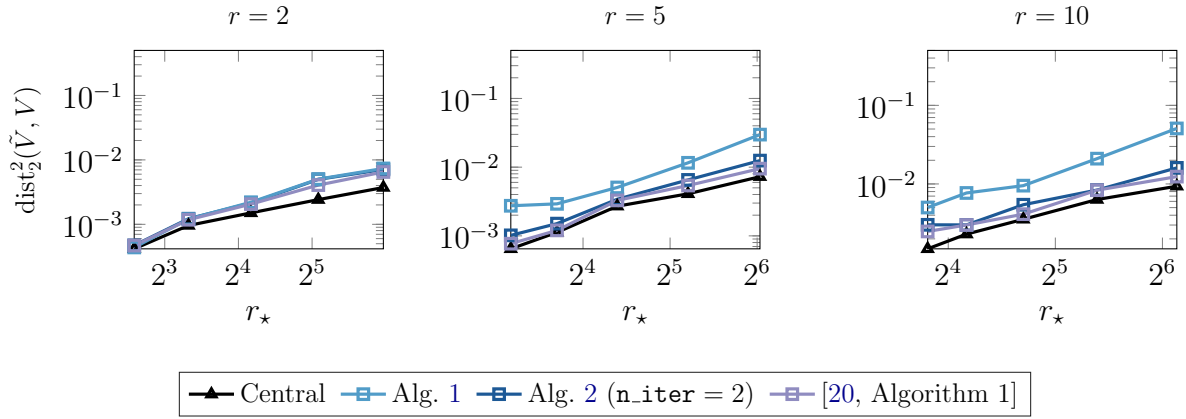
\begin{figure}[tb]
    \centering
    \begin{minipage}{0.325\textwidth}
        \centering
        \begin{tikzpicture}
            \begin{axis}[xlabel=$r_{\star}$,ylabel=\scalebox{0.85}{${\dist_2^2(\tilde{V}, V)}$},
                title={\scalebox{0.85}{$r=2$}},
                enlargelimits=false, width=0.93\linewidth, legend cell align=left,
                ymode=log, ymax=0.5, legend style={nodes={scale=0.75, transform shape}}, xmode=log, log basis x={2}]
            \addplot[mark=triangle, very thick, color=black]
                table[x=srs, y=erm, col sep=comma] {fig_data/error-2_maxr-64_n_iter-2.csv};
            \addplot[mark=square, very thick, color=hblue]
                table[x=srs, y=itr, col sep=comma] {fig_data/error-2_maxr-64_n_iter-2.csv};
            \addplot[mark=square, very thick, color=mblue]
                table[x=srs, y=fix, col sep=comma] {fig_data/error-2_maxr-64_n_iter-2.csv};
            \addplot[mark=square, very thick, color=lblue]
                table[x=srs, y=rot, col sep=comma] {fig_data/error-2_maxr-64_n_iter-2.csv};
            \end{axis}
        \end{tikzpicture}
    \end{minipage}~
    \begin{minipage}{0.325\textwidth}
        \centering
        \begin{tikzpicture}
            \begin{axis}[xlabel=$r_{\star}$, title={\scalebox{0.85}{$r=5$}},
            enlargelimits=false, width=0.93\linewidth, legend cell align=left, ymode=log, ymax=0.5,
            legend style={nodes={scale=0.75, transform shape}}, xmode=log, log basis x={2}]
            \addplot[mark=triangle, very thick, color=black]
                table[x=srs, y=erm, col sep=comma] {fig_data/error-5_maxr-64_n_iter-2.csv};
            \addplot[mark=square, very thick, color=hblue]
                table[x=srs, y=itr, col sep=comma] {fig_data/error-5_maxr-64_n_iter-2.csv};
            \addplot[mark=square, very thick, color=mblue]
                table[x=srs, y=fix, col sep=comma] {fig_data/error-5_maxr-64_n_iter-2.csv};
            \addplot[mark=square, very thick, color=lblue]
                table[x=srs, y=rot, col sep=comma] {fig_data/error-5_maxr-64_n_iter-2.csv};
            \end{axis}
        \end{tikzpicture}
    \end{minipage}~
    \begin{minipage}{0.325\textwidth}
        \centering
        \begin{tikzpicture}
            \begin{axis}[xlabel=$r_{\star}$, title={\scalebox{0.85}{$r=10$}},
            enlargelimits=false, width=0.93\linewidth, legend cell align=left, ymode=log, ymax=0.5,
            legend style={nodes={scale=0.75, transform shape}}, xmode=log, log basis x={2}]
            \addplot[mark=triangle, very thick, color=black]
                table[x=srs, y=erm, col sep=comma] {fig_data/error-10_maxr-64_n_iter-2.csv};
            \addplot[mark=square, very thick, color=hblue]
                table[x=srs, y=itr, col sep=comma] {fig_data/error-10_maxr-64_n_iter-2.csv};
            \addplot[mark=square, very thick, color=mblue]
                table[x=srs, y=fix, col sep=comma] {fig_data/error-10_maxr-64_n_iter-2.csv};
            \addplot[mark=square, very thick, color=lblue]
                table[x=srs, y=rot, col sep=comma] {fig_data/error-10_maxr-64_n_iter-2.csv};
            \end{axis}
        \end{tikzpicture}
    \end{minipage}\\
    \begin{center}
    \begin{tikzpicture}
        \begin{axis}[%
        hide axis,
        xmin=10,
        xmax=50,
        ymin=0,
        ymax=0.4,
        legend style={draw=white!15!black,legend cell align=left},
        legend columns=4,
        legend style={nodes={scale=0.85, transform shape}}
        ]
        \addlegendimage{mark=triangle, color=black, very thick};
        \addlegendentry{Central \qquad};
        \addlegendimage{mark=square, color=mblue, very thick};
        \addlegendentry{Alg.~\ref{alg:procrustes-fixing} \qquad};
        \addlegendimage{mark=square, color=hblue, very thick};
        \addlegendentry{Alg.~\ref{alg:procrustes-fixing-iterative} ($\texttt{n\_iter} = 2$) \qquad};
        \addlegendimage{mark=square, very thick, color=lblue};
        \addlegendentry{\cite[Algorithm 1]{FWWZ19}};
        \end{axis}
    \end{tikzpicture}
    \end{center}
    \caption{Performance of~\cref{alg:procrustes-fixing,alg:procrustes-fixing-iterative}
        compared to centralized PCA and~\cite[Algorithm 1]{FWWZ19}, for $d = 250, n = 500,
        m = 100$, given a covariance matrix $X$ of varying intrinsic dimension $r_{\star} =
        \textsf{intdim}(X)$. Here, $\delta = 0.25$
        and $r \in \set{2, 5, 10}$ (\textbf{left to right}).
        In all cases, the errors of~\cref{alg:procrustes-fixing,alg:procrustes-fixing-iterative}
        are at worst within a constant factor of the error of the centralized algorithm.}
    \label{fig:lowrank-numerics}
\end{figure}

To examine the influence of $r_{\star} = \mathsf{intdim}(X)$ on the performance of the
algorithm, we generate a family of covariance matrices following the
model~\eqref{eq:intdim-experiment} by letting
\[
    r_{\star} \in \set{r + 2^{k} \mmid k = 2, \dots, 6},
\]
where $r$ is the dimension of the principal subspace, while keeping $d = 250,
n = 2 \cdot d$, and $m = 100$ fixed. \Cref{fig:lowrank-numerics}
depicts the results for Gaussian samples generated according to~\cref{eq:intdim-experiment}
and $r \in \set{2, 5, 10}$.  In all cases, the performance of~\cref{alg:procrustes-fixing}
is competitive to centralized PCA, as well as with~\cite[Algorithm 1]{FWWZ19},
its error being at worst a constant multiplicative factor larger than
centralized PCA. In addition, the performance of \cref{alg:procrustes-fixing-iterative},
the iteratively refined variant of \cref{alg:procrustes-fixing} introduced
in~\Cref{sec:iterative-refinement}, is on par with both centralized PCA as well
as~\cite[Algorithm 1]{FWWZ19}. \newstuff{In all cases, an increase in the intrinsic dimension
$r_{\star}$ implies an increase in the error of all estimators.}

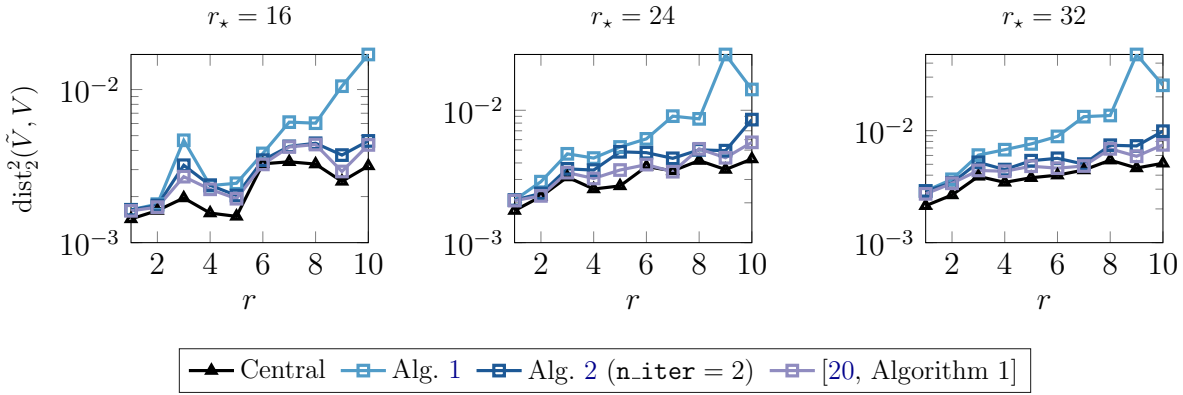
\begin{figure}[tb]
    \centering
    \begin{minipage}{0.325\textwidth}
        \centering
        \begin{tikzpicture}
            \begin{axis}[xlabel=$r$, ylabel=\scalebox{0.85}{${\dist_2^2(\tilde{V}, V)}$},
                title={\scalebox{0.85}{$r_{\star}=16$}},
                ymin=1e-3,
                enlargelimits=false, width=0.9\linewidth, ymode=log]
            \addplot[mark=triangle, very thick, color=black]
                table[x=r, y=erm, col sep=comma] {fig_data/error_stablerank-16_reps-5.csv};
            \addplot[mark=square, very thick, color=mblue]
                table[x=r, y=one, col sep=comma] {fig_data/error_stablerank-16_reps-5.csv};
            \addplot[mark=square, very thick, color=hblue]
                table[x=r, y=fix, col sep=comma] {fig_data/error_stablerank-16_reps-5.csv};
            \addplot[mark=square, very thick, color=lblue]
                table[x=r, y=rot, col sep=comma] {fig_data/error_stablerank-16_reps-5.csv};
            \end{axis}
        \end{tikzpicture}
    \end{minipage}~
    \begin{minipage}{0.325\textwidth}
        \centering
        \begin{tikzpicture}
            \begin{axis}[
                xlabel=$r$, title={\scalebox{0.85}{$r_{\star}=24$}},
                ymin=1e-3,
                enlargelimits=false, ymode=log, width=0.9\linewidth]
            \addplot[mark=triangle, very thick, color=black]
                table[x=r, y=erm, col sep=comma] {fig_data/error_stablerank-24_reps-5.csv};
            \addplot[mark=square, very thick, color=mblue]
                table[x=r, y=one, col sep=comma] {fig_data/error_stablerank-24_reps-5.csv};
            \addplot[mark=square, very thick, color=hblue]
                table[x=r, y=fix, col sep=comma] {fig_data/error_stablerank-24_reps-5.csv};
            \addplot[mark=square, very thick, color=lblue]
                table[x=r, y=rot, col sep=comma] {fig_data/error_stablerank-24_reps-5.csv};
            \end{axis}
        \end{tikzpicture}
    \end{minipage}~
    \begin{minipage}{0.325\textwidth}
        \centering
        \begin{tikzpicture}
            \begin{axis}[xlabel=$r$,
                title={\scalebox{0.85}{$r_{\star}=32$}},
                ymin=1e-3,
                enlargelimits=false, width=0.9\linewidth, ymode=log]
            \addplot[mark=triangle, very thick, color=black]
                table[x=r, y=erm, col sep=comma] {fig_data/error_stablerank-32_reps-5.csv};
            \addplot[mark=square, very thick, color=mblue]
                table[x=r, y=one, col sep=comma] {fig_data/error_stablerank-32_reps-5.csv};
            \addplot[mark=square, very thick, color=hblue]
                table[x=r, y=fix, col sep=comma] {fig_data/error_stablerank-32_reps-5.csv};
            \addplot[mark=square, very thick, color=lblue]
                table[x=r, y=rot, col sep=comma] {fig_data/error_stablerank-32_reps-5.csv};
            \end{axis}
        \end{tikzpicture}
    \end{minipage}~
    \begin{center}
    \begin{tikzpicture}
        \begin{axis}[%
        hide axis,
        xmin=10,
        xmax=50,
        ymin=0,
        ymax=0.4,
        legend style={draw=white!15!black,legend cell align=left},
        legend columns=4,
        legend style={nodes={scale=0.85, transform shape}}
        ]
        \addlegendimage{mark=triangle, color=black, very thick};
        \addlegendentry{Central \qquad};
        \addlegendimage{mark=square, color=mblue, very thick};
        \addlegendentry{Alg.~\ref{alg:procrustes-fixing} \qquad};
        \addlegendimage{mark=square, color=hblue, very thick};
        \addlegendentry{Alg.~\ref{alg:procrustes-fixing-iterative} ($\texttt{n\_iter} = 2$) \qquad};
        \addlegendimage{mark=square, very thick, color=lblue};
        \addlegendentry{\cite[Algorithm 1]{FWWZ19}};
        \end{axis}
    \end{tikzpicture}
    \end{center}
    \caption{Performance of~\cref{alg:procrustes-fixing,alg:procrustes-fixing-iterative}
        compared to centralized PCA and~\cite[Algorithm 1]{FWWZ19}, for $d = 250, n = 500, m = 100$,
        for varying ranks $r$ over instances of fixed intrinsic dimension.
        Here, $\delta = 0.25$ and $r_{\star} \in \set{16, 24, 32}$ (\textbf{left to right}).
        In all cases, the errors of~\cref{alg:procrustes-fixing,alg:procrustes-fixing-iterative}
        are at worst within a constant factor of the error of the centralized algorithm.}
        \label{fig:fixedrank-numerics}
\end{figure}

\newstuff{Moreover, we set up an additional experiment where the intrinsic dimension
$r_{\star}$ was kept fixed while varying the target subspace dimension $r$,
for the same choice of parameters $d = 250, n = 2\cdot d, m = 100$ and eigengap
equal to $\delta = 0.25$, using Gaussian samples. In particular, we tried $3$
sets of instances where $r_{\star} \in \set{16, 24, 32}$ and $r$ is varied
between $1$ and $10$, and compared~\cref{alg:procrustes-fixing,alg:procrustes-fixing-iterative}
with the centralized estimator as well as~\cite[Algorithm 1]{FWWZ19}.
The experimental findings are depicted in~\cref{fig:fixedrank-numerics}; we
observe that the error follows an increasing trend as $r$ varies.
However, this does not contradict our theoretical results for the
following reasons:
\begin{itemize}
    \item Despite the increasing trend, there are cases where an increase in
        $r$ does not lead to an increase in error, in contrast to~\cref{fig:lowrank-numerics}.
    \item The centralized estimator, whose theoretical performance depends on $r_{\star}$
        (see, e.g.,~\cite[Theorem 9.2.4]{Vershynin18}), follows the same trend.
        Indeed, the dependence on $r_{\star}$ characterizes the worst-case error of the
        estimator as quantified by concentration bounds, while the per-instance dependence
        on $r$ and $r_{\star}$ may be more nuanced.
\end{itemize}
}

\subsection{Experiments with non-gaussian measurements}
So far, we have noticed that~\cref{alg:procrustes-fixing} achieves comparable
yet lower accuracy than Algorithm 1 in~\cite{FWWZ19};~\cref{alg:procrustes-fixing-iterative}
closes this gap but stil does not achieve strictly better estimation error. An
interpretation of this is due to the fact that the Procrustes-based algorithm
always introduces some bias to the resulting estimator, whereas the algorithm
of~\cite{FWWZ19} is unbiased for Gaussian data (which has been the focus of our
synthetic experiments so far).

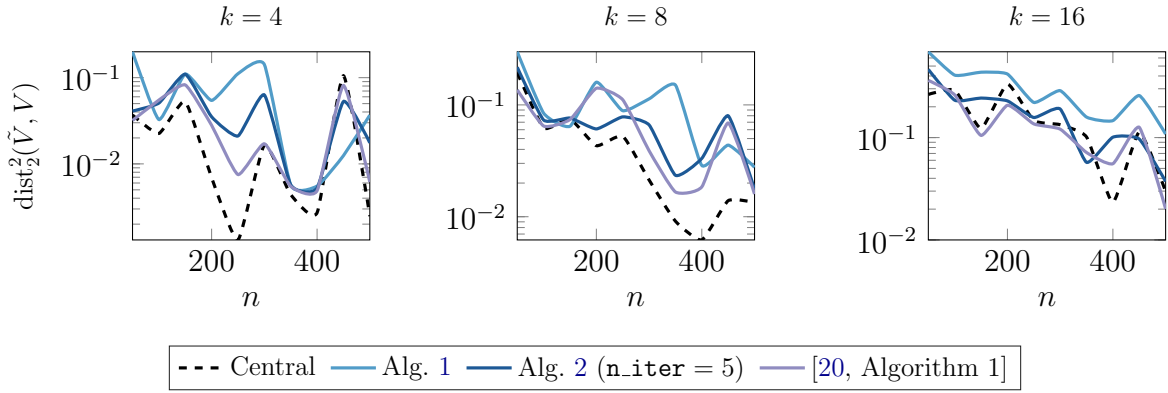
\begin{figure}[tb]
    \centering
    \begin{minipage}{0.325\textwidth}
        \centering
        \begin{tikzpicture}
            \begin{axis}[xlabel=$n$, ylabel=\scalebox{0.85}{${\dist_2^2(\tilde{V}, V)}$},
                title={\scalebox{0.85}{$k=4$}},
                enlargelimits=false, width=0.9\linewidth, ymode=log,
                tension=0.3]
            \addplot[no markers, dashed, smooth, very thick, color=black]
                table[x=n, y=erm, col sep=comma] {fig_data/error_uniform-4_m-25_reps-10.csv};
            \addplot[no markers, smooth, very thick, color=mblue]
                table[x=n, y=one, col sep=comma] {fig_data/error_uniform-4_m-25_reps-10.csv};
            \addplot[no markers, smooth, very thick, color=hblue]
                table[x=n, y=fix, col sep=comma] {fig_data/error_uniform-4_m-25_reps-10.csv};
            \addplot[no markers, smooth, very thick, color=lblue]
                table[x=n, y=rot, col sep=comma] {fig_data/error_uniform-4_m-25_reps-10.csv};
            \end{axis}
        \end{tikzpicture}
    \end{minipage}~
    \begin{minipage}{0.325\textwidth}
        \centering
        \begin{tikzpicture}
            \begin{axis}[
                xlabel=$n$, title={\scalebox{0.85}{$k = 8$}},
                enlargelimits=false, ymode=log, width=0.9\linewidth,
                tension=0.3]
            \addplot[no markers, dashed, very thick, smooth, color=black, mark options={solid}]
                table[x=n, y=erm, col sep=comma] {fig_data/error_uniform-8_m-25_reps-10.csv};
            \addplot[no markers, smooth, very thick, color=mblue]
                table[x=n, y=one, col sep=comma] {fig_data/error_uniform-8_m-25_reps-10.csv};
            \addplot[no markers, smooth, very thick, color=hblue]
                table[x=n, y=fix, col sep=comma] {fig_data/error_uniform-8_m-25_reps-10.csv};
            \addplot[no markers, smooth, very thick, color=lblue]
                table[x=n, y=rot, col sep=comma] {fig_data/error_uniform-8_m-25_reps-10.csv};
            \end{axis}
        \end{tikzpicture}
    \end{minipage}~
    \begin{minipage}{0.325\textwidth}
        \centering
        \begin{tikzpicture}
            \begin{axis}[xlabel=$n$,
                title={\scalebox{0.85}{$k=16$}},
                enlargelimits=false, width=0.9\linewidth, ymode=log, ymin=1e-2,
                tension=0.3]
            \addplot[no markers, smooth, dashed, very thick, color=black]
                table[x=n, y=erm, col sep=comma] {fig_data/error_uniform-16_m-25_reps-10.csv};
            \addplot[no markers, smooth, very thick, color=mblue]
                table[x=n, y=one, col sep=comma] {fig_data/error_uniform-16_m-25_reps-10.csv};
            \addplot[no markers, smooth, very thick, color=hblue]
                table[x=n, y=fix, col sep=comma] {fig_data/error_uniform-16_m-25_reps-10.csv};
            \addplot[no markers, smooth, very thick, color=lblue]
                table[x=n, y=rot, col sep=comma] {fig_data/error_uniform-16_m-25_reps-10.csv};
            \end{axis}
        \end{tikzpicture}
    \end{minipage}~
    \begin{center}
    \begin{tikzpicture}
        \begin{axis}[%
        hide axis,
        xmin=10,
        xmax=50,
        ymin=0,
        ymax=0.4,
        legend style={draw=white!15!black,legend cell align=left},
        legend columns=4,
        legend style={nodes={scale=0.85, transform shape}}
        ]
        \addlegendimage{no markers, color=black, dashed, very thick};
        \addlegendentry{Central \qquad};
        \addlegendimage{no markers, color=mblue, very thick};
        \addlegendentry{Alg.~\ref{alg:procrustes-fixing} \qquad};
        \addlegendimage{no markers, color=hblue, very thick};
        \addlegendentry{Alg.~\ref{alg:procrustes-fixing-iterative} ($\texttt{n\_iter} = 5$) \qquad};
        \addlegendimage{no markers, very thick, color=lblue};
        \addlegendentry{\cite[Algorithm 1]{FWWZ19}};
        \end{axis}
    \end{tikzpicture}
    \end{center}
    \caption{Performance of~\cref{alg:procrustes-fixing,alg:procrustes-fixing-iterative}
        compared to centralized PCA and~\cite[Algorithm 1]{FWWZ19} for samples drawn from
        $\cD_k$ in~\eqref{eq:discrete-uniform} for varying $k \in \set{4, 8, 16}$ (\textbf{left to right}).
        In all cases, we compute the leading eigenspace of dimension $r = \frac{k}{2}$.
        Note that Alg. 1 from~\cite{FWWZ19} achieves the lowest error in most, but not all,
    instances.}
    \label{fig:discrete-uniform}
\end{figure}

Here, we show that this phenomenon persists empirically even when moving to non-Gaussian
data. In particular, we generate our samples from the following distribution:
\begin{align}
    \cD_k = \mathrm{Unif}\set{y_1, \dots, y_k}, \quad \text{where} \quad
    y_i \in \sqrt{d} \Sbb^{d-1}.
    \label{eq:discrete-uniform}
\end{align}
The distribution $\cD_k$ is not Gaussian and, by the remarks in~\cite[Section 5.6]{Vershynin18},
is heavy-tailed unless the number of vectors $k$ grows exponentially in $d$. To
avoid having to deal with centering issues, we estimate the leading eigenspace of
the second moment matrix (instead of the covariance matrix) of $\cD_{k}$.
\Cref{fig:discrete-uniform} depicts the results for an experiment with $m = 25$ machines, number
of samples $n \in \set{50, 100, \dots, 500}$, and $k \in \set{4, 8, 16}$. Remarkably,
we find that Alg. 1 from~\cite{FWWZ19} results in lower estimation error in most, but
not all, instances generated.

\subsection{Consistence of theoretical predictions}
In this section, we examine how the empirical error compares with the theoretically
prescribed rate from~\cref{theorem:stablerank-rate}, which provides a refined bound
using the intrinsic dimension.
We compare $\dist_2(V, \tilde{V})$ with the following (simplified) bound
$f(r_{\star}, n)$, for various configurations of $r_{\star}, n$ and
fixing $(d, m) = (300, 100), \; \delta = 0.2$:
\begin{equation}
    f(r_{\star}, n) := \frac{r_{\star} + \log m}{\delta^2 n}
    + \sqrt{\frac{r_{\star} + 2 \log(n)}{\delta^2 mn}}.
    \label{eq:theo-rate-simplified}
\end{equation}
Notice that, due to our construction of the eigenvalues of the covariance
matrix shown in~\cref{eq:covmat-construction}, higher values for $\dim(V_1)$
increase $\textsf{intdim}(\Sigma)$.
In the figures below, we report the median over $10$ independent runs of the
algorithm.

\begin{figure}[tb]
    \centering
    \begin{minipage}{0.55\textwidth}
    \begin{tikzpicture}
        \begin{axis}[xlabel=$n$,ylabel=${\dist_2(\tilde{V}, V)}$,
                    enlargelimits=false, width=0.85\linewidth, legend cell align=left,
                    xmode=log, log basis x={2},
                    ymode=log, legend style={nodes={scale=0.8, transform shape}},
                    ymax=5.0, mark options={solid},
                    legend entries={
                        {$r = 5$}, {$r = 10$}, {$r = 20$},
                        {$f(n)$}, {Alg.~\ref{alg:procrustes-fixing}}}]
            \addlegendimage{no markers, very thick, color=black};
            \addlegendimage{no markers, very thick, color=hblue};
            \addlegendimage{no markers, very thick, color=lblue}
            \addlegendimage{only marks, mark=square, black!75, very thick};
            \addlegendimage{only marks, mark=triangle, black!75, very thick};
            \addplot[mark=square, dashed, very thick, color=black]
                table[x=n, y=theo, col sep=comma] {fig_data/error_theory-5_gap-0.20.csv};
            \addplot[mark=square, dashed, very thick, color=hblue]
                table[x=n, y=theo, col sep=comma] {fig_data/error_theory-10_gap-0.20.csv};
            \addplot[mark=square, dashed, very thick, color=lblue]
                table[x=n, y=theo, col sep=comma] {fig_data/error_theory-20_gap-0.20.csv};
            \addplot[mark=triangle, dashed, very thick, color=black]
                table[x=n, y=fix, col sep=comma] {fig_data/error_theory-5_gap-0.20.csv};
            \addplot[mark=triangle, dashed, very thick, color=hblue]
                table[x=n, y=fix, col sep=comma] {fig_data/error_theory-10_gap-0.20.csv};
            \addplot[mark=triangle, dashed, very thick, color=lblue]
                table[x=n, y=fix, col sep=comma] {fig_data/error_theory-20_gap-0.20.csv};
        \end{axis}
    \end{tikzpicture}
    \end{minipage}
    \caption{Empirical error of~\cref{alg:procrustes-fixing} vs.\ theoretically
        prescribed error $f(r_{\star}, n)$ from~\cref{theorem:stablerank-rate}, the
    latter of which is loose by an order of magnitude. Here, $(d, m) = (300, 100)$ and $\delta = 0.2$.}
    \label{fig:theory}
\end{figure}
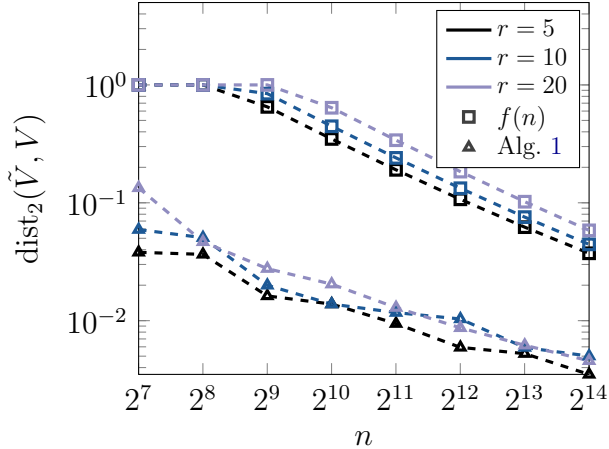

\Cref{fig:theory} reveals that the empirical error rate of
Algorithm~\ref{alg:procrustes-fixing} is well below the theoretically prescribed
rate, as $f(r_{\star}, n)$ is an order of magnitude loose compared to the subspace
distance $\dist_2(\tilde{V}, V_1)$.
This is not too surprising,
since some of the intermediate results used to arrive at~\cref{proposition:average-subspace}
depend on worst-case perturbation bounds which may not materialize in practice.

\subsection{Application: distributed node embeddings}
We apply our algorithm to the task of generating \textit{node embeddings}
for undirected graphs. Given a graph $G = (V, E)$, methods for generating node
embeddings generate a $d$ -- dimensional vector ${z}_i \in \Rbb^d$ for all nodes
$v_i \in V$. One broad category of unsupervised methods for this task is
so-called \textit{implicit factorization methods}~\cite{HYL17}, which
attempt to directly minimize a loss of the form
\begin{equation}
    \sum_{(v_i, v_j) \in \cD} \norm{\ip{z_i, z_j}
    - s_{G}(v_i, v_j)}^2
    \approx \norm{ZZ^\T - S_G}_F^2,
    \label{eq:node-embed-mf}
\end{equation}
where $\cD$ is a set of node pairs (e.g. all connected nodes) and
$s_G(v_i, v_j)$ is a proximity measure between nodes $v_i, v_j$ (for
example, $s_G(v_i, v_j) = \bm{1}\set{(v_i, v_j) \in E}$). Since the loss
is invariant to arbitrary orthogonal transforms of the node embeddings,
our algorithmic framework is also applicable to combining node embeddings
in a distributed environment, in the setting described below.

\paragraph{Experimental setup}
Consider a graph $G = (V, E)$, and let $m$ denote the number
of machines. We assume that machine $i$ observes a ``censored'' version
$G^{(i)} = (V, E^{(i)})$ of the graph $G$, where edges are ``hidden''
independently with probability $p$, so
that $\Pbb(E^{(i)}_{j, k} = 1) = (1 - p) \cdot \bm{1}\set{E_{j, k} = 1}$, leading to
$\expec{E^{(i)}} = (1 - p) E, \; \forall i$. Each machine then applies a node
embedding algorithm to $G^i$. In our experiments, we set the edge failure
probability $p = 0.1$ and use the \texttt{HOPE}
method~\cite{OCP+16} with embedding dimension $d = 64$ and path decay $\beta = 0.1$.

Letting ${Z}^{(i)} \in \Rbb^{\abs{V} \times d}$ denote the matrix of node
embeddings for the $i^{\text{th}}$ compute node, we define the following
solutions:
\begin{itemize}
    \item ${Z}_{\text{avg}} := \frac{1}{m} \sum_{i=1}^m {Z}^{(i)}
        Q^{(i)}$ for the Procrustes-aligned estimate, where
        \[
            Q^{(i)} := \argmin_{U \in \Obb_d} \norm{{Z}^{(i)} U - {Z}^{(1)}}_F.
        \]
    \item ${Z}_{\text{cnt}}$ for the ``central'' estimate, which is the node
        embedding generated applying the method on the ``uncensored'' version
        of $G$.
    \item $Z_{\text{naive}}$, for the vanilla averaged estimate $\frac{1}{m}
        \sum_{i} Z^{(i)}$.
\end{itemize}
We evaluate the solutions in a node classification setting based on Macro-F1 score,
using the \texttt{Wikipedia}~\cite{Wikipedia}
and \texttt{PPI} (Protein-Protein Interaction)~\cite{BioGRID} datasets.
The node embeddings serve as features for a logistic regression classifier
(after standardization) with $\ell_2$ regularization of inverse strength
$C = 0.5$ (\texttt{Wikipedia}) and $C = 1.0$ (\texttt{PPI}). We use a
$75\%$ / $25\%$ split of training / test data. Finally, we average classification
metrics over $10$ random instantiations of such splits.
\Cref{table:F1-scores} depicts the relative loss in F1 score when using $Z_{\text{avg}}$ instead of
$Z_{\text{cnt}}$; we see that the relative loss is minimal (with the exception
of a single configuration), with several configurations actually benefit from
using the averaged solution.

\begin{figure}[tb]
    \centering
    \begin{minipage}{0.42\textwidth}
        \begin{tikzpicture}
            \begin{axis}[xlabel=$m$,ylabel={$\dist_{\norm{\cdot}_F}(\cdot, Z_{\text{cnt}})$},
                        width=0.95\linewidth, enlargelimits=false,
                        legend cell align=left, title={\texttt{Wikipedia}},
                        legend style={nodes={scale=0.8, transform shape},
                        at={(0.97, 0.5)}, anchor=east},
                        mark options={solid}, xmode=log, log basis x={2}]
                \addplot[mark=square, solid, very thick, color=hblue]
                    table[x=numNodes, y=avgDists, col sep=comma] {fig_data/pos_hope_0.75.csv};
                \addplot[mark=*, dashed, very thick, color=black]
                    table[x=numNodes, y=nveDists, col sep=comma] {fig_data/pos_hope_0.75.csv};
                \legend{{$Z_{\text{avg}}$}, {$Z_{\text{naive}}$}};
            \end{axis}
        \end{tikzpicture}
    \end{minipage}~
    \begin{minipage}{0.42\textwidth}
        \begin{tikzpicture}
            \begin{axis}[xlabel=$m$,ylabel={$\dist_{\norm{\cdot}_F}(\cdot, Z_{\text{cnt}})$},
                        width=0.95\linewidth, enlargelimits=false,
                        legend cell align=left, title={\texttt{PPI}},
                        legend style={nodes={scale=0.8, transform shape},
                        at={(0.97, 0.5)}, anchor=east},
                        mark options={solid}, xmode=log, log basis x={2}]
                \addplot[mark=square, solid, very thick, color=hblue]
                    table[x=numNodes, y=avgDists, col sep=comma] {fig_data/ppi_hope_0.75.csv};
                \addplot[mark=*, dashed, very thick, color=black]
                    table[x=numNodes, y=nveDists, col sep=comma] {fig_data/ppi_hope_0.75.csv};
                \legend{{$Z_{\text{avg}}$}, {$Z_{\text{naive}}$}};
            \end{axis}
        \end{tikzpicture}
    \end{minipage}
    \caption{Distance of solutions produced by naive averaging and~\cref{alg:procrustes-fixing}
    from centralized solution for the \texttt{Wikipedia} and \texttt{PPI} datasets.
    The quality of the solution constructed by~\cref{alg:procrustes-fixing} does not
    degrade with $m$, as opposed to naive averaging.}
    \label{fig:hope-distances}
\end{figure}
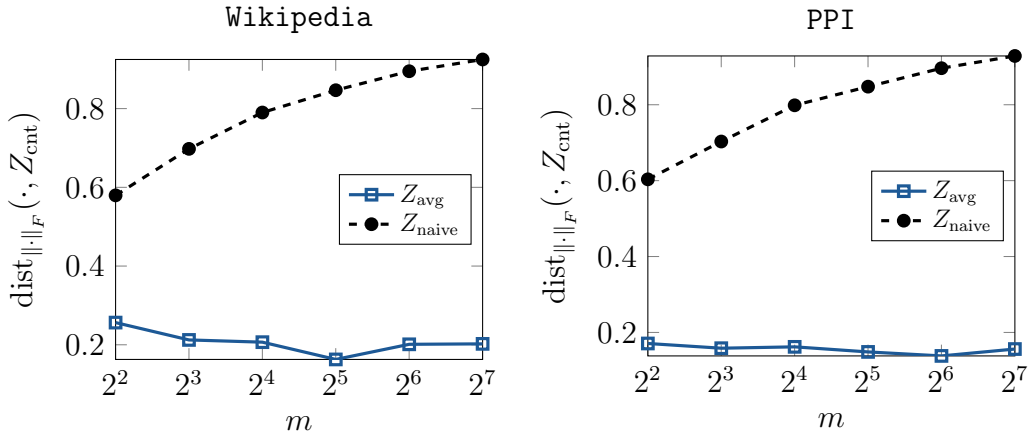

In addition, we compare the distances of $Z_{\text{avg}}$ and
$Z_{\text{naive}}$ from the ``central''
embedding $Z_{\text{cnt}}$, as depicted in~\cref{fig:hope-distances}.
Unsurprisingly, as the number of machines $m$ grows, $Z_{\text{naive}}$ strays
further away from the embedding generated from a consistent view of $G$. On the
other hand, averaging using~\cref{alg:procrustes-fixing} leads to estimates whose
distance from the centralized solution does not increase with $m$.
{
    \renewcommand{\arraystretch}{1.05}
    \aboverulesep=0ex
    \belowrulesep=0ex
    \begin{table}[tb]
        \centering
        \caption{Relative decrease in macro-F1 score using $Z_{\text{avg}}$ (\cref{alg:procrustes-fixing} averaging on several censored graphs) instead of
          $Z_{\text{cnt}}$ (central solution on entire graph) for one vs.\ rest logistic regression. In most cases, the distributed
          solution does not reduce the predictive performance (in several occasions, using
          the averaged solution actually leads to a better macro-F1 score).}
        \begin{tabular}{r @{\qquad} r c c c c c c}
          \toprule
          \textbf{Dataset} & $m =$ & $2^2$ & $2^3$ & $2^4$ & $2^5$ & $2^6$ & $2^7$ \\ \midrule
          \texttt{Wikipedia} & & 0.0 & 0.0 & 0.84\% & 0.0 & 0.0 & 0.38\% \\
          \texttt{PPI} & & 27.01\% & 0.0 & 0.0 & 0.0 & 0.0 & 0.0 \\
          \bottomrule
        \end{tabular}
        \label{table:F1-scores}
    \end{table}
}

\subsection{Application: distributed spectral initializations}
We present an application of our algorithm to a distributed method for
initializing local search algorithms for \textbf{quadratic sensing}.
In this
setting, one observes $N$ measurements of the form
\begin{equation}
    y_i = \norm{X_{\sharp}^\T a_i}_2^2 + \texttt{noise}_i, \quad i \in [N]
\end{equation}
where $\set{a_i}_{i \in [N]}$ are \textit{known} design vectors, typically
satisfying $a_i \iid \cN(0, I_d)$, and $X_{\sharp} \in \Rbb^{d \times r}$ is a
matrix to be recovered. Applications of quadratic sensing include covariance
sketching~\cite{CCG15}, learning one-hidden-layer networks with quadratic
activations~\cite{LSS+14}, and quantum state tomography~\cite{KRT17}. When
$r = 1$, quadratic sensing recovers the well-known
\textbf{phase retrieval} problem, which can be viewed as in instance of the
Gaussian \textit{single-index model}, also discussed in the applications
of~\cite{CLLY20}.
Even though the natural least-squares formulation of this problem is nonconvex,
there are provable algorithms in the literature (see, e.g.,~\cite{CLS15,CC15,CLC19}
for a non-exhaustive list) that retrieve a global solution (up to rotation) by
performing a carefully initialized local search. Assuming for simplicity that
$X_{\sharp} \in \Obb_{d, r}$, a natural spectral initialization for such
methods works as follows: we form the positive semidefinite matrix $D_N$, given by
\begin{equation}
    D_N := \frac{1}{N} \sum_{i=1}^N \cT(y_i) a_i a_i^\T,
    \quad \cT(\cdot) : \Rbb \to \Rbb_+,
    \label{eq:D_m}
\end{equation}
where $\cT$ is commonly chosen as a truncation operator, e.g. $\cT(y) :=
y \1\set{y \leq \tau}$ for some threshold $\tau$. Standard arguments then
show that the leading $r$-dimensional eigenspace (denoted by $X_0$) of $D_N$
forms a nontrivial estimate of $X_{\sharp}$ with high probability, as long as
the number of samples $N \gtrsim r^4 d \log(d)$~\cite{CLS15}.

Our alignment framework can be applied to the problem at hand when the measurements
$y_i$ are collected in several different machines.
To initialize, e.g., stochastic local
search algorithms, each machine can form its ``local'' $D_N$ matrix and compute
a weak estimate $X_0$ of $X_{\sharp}$; a central coordinator can then refine the
weak estimate by aggregation and redistribution to local machines. Even
though our theory is not directly transferrable, we nevertheless demonstrate that
Procrustes-fixing can be applied for this more general problem.

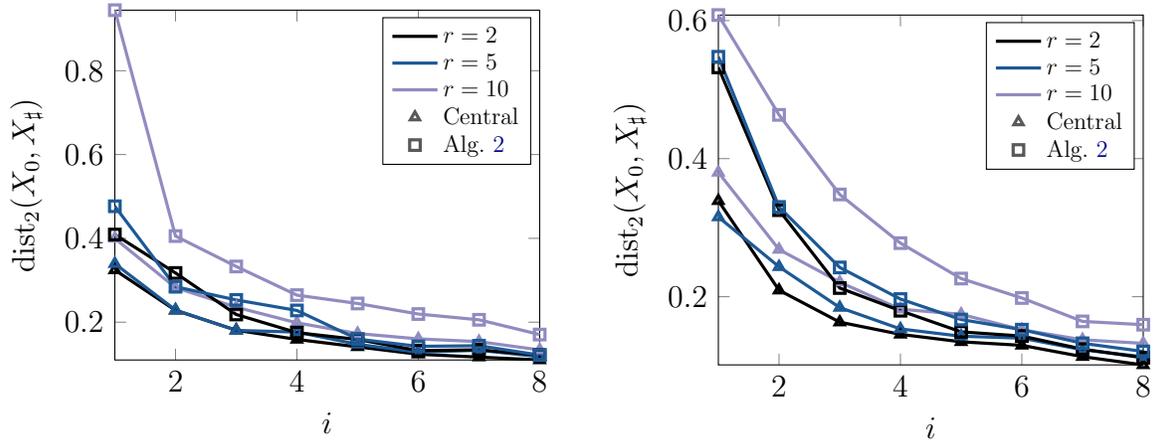
\begin{figure}[tb]
    \centering
    \begin{minipage}{0.47\textwidth}
    \begin{tikzpicture}
        \begin{axis}[xlabel=$i$, ylabel=${\dist_2(X_0, X_{\sharp})}$,
                    enlargelimits=false, width=0.95\linewidth, legend cell align=left,
                    legend style={nodes={scale=0.75, transform shape}},
                    mark options={solid}, 
                    legend entries={
                        {$r = 2$}, {$r = 5$}, {$r = 10$},
                    {Central}, {Alg.~\ref{alg:procrustes-fixing-iterative}}}]
            \addlegendimage{no markers, very thick, color=black};
            \addlegendimage{no markers, very thick, color=hblue};
            \addlegendimage{no markers, very thick, color=lblue}
            \addlegendimage{only marks, mark=triangle, black!65, very thick};
            \addlegendimage{only marks, mark=square, black!65, very thick};
            \addplot[mark=triangle, very thick, color=black]
                table[x=i, y=erm, col sep=comma] {fig_data/error_quadsensing_100x2_n_iter-10.csv};
            \addplot[mark=triangle, very thick, color=hblue]
                table[x=i, y=erm, col sep=comma] {fig_data/error_quadsensing_100x5_n_iter-10.csv};
            \addplot[mark=triangle, very thick, color=lblue]
                table[x=i, y=erm, col sep=comma] {fig_data/error_quadsensing_100x10_n_iter-10.csv};
            \addplot[mark=square, very thick, color=black]
                table[x=i, y=fix, col sep=comma] {fig_data/error_quadsensing_100x2_n_iter-10.csv};
            \addplot[mark=square, very thick, color=hblue]
                table[x=i, y=fix, col sep=comma] {fig_data/error_quadsensing_100x5_n_iter-10.csv};
            \addplot[mark=square, very thick, color=lblue]
                table[x=i, y=fix, col sep=comma] {fig_data/error_quadsensing_100x10_n_iter-10.csv};
        \end{axis}
    \end{tikzpicture}
    \end{minipage}\quad
    \begin{minipage}{0.47\textwidth}
    \begin{tikzpicture}
        \begin{axis}[xlabel=$i$, ylabel=${\dist_2(X_0, X_{\sharp})}$,
                    enlargelimits=false, width=0.95\linewidth, legend cell align=left,
                    legend style={nodes={scale=0.75, transform shape}},
                    mark options={solid}, 
                    legend entries={
                        {$r = 2$}, {$r = 5$}, {$r = 10$},
                    {Central},
                {Alg.~\ref{alg:procrustes-fixing-iterative}}}]
            \addlegendimage{no markers, very thick, color=black};
            \addlegendimage{no markers, very thick, color=hblue};
            \addlegendimage{no markers, very thick, color=lblue}
            \addlegendimage{only marks, mark=triangle, black!65, very thick};
            \addlegendimage{only marks, mark=square, black!65, very thick};
            \addplot[mark=triangle, very thick, color=black]
                table[x=i, y=erm, col sep=comma] {fig_data/error_quadsensing_200x2_n_iter-10.csv};
            \addplot[mark=triangle, very thick, color=hblue]
                table[x=i, y=erm, col sep=comma] {fig_data/error_quadsensing_200x5_n_iter-10.csv};
            \addplot[mark=triangle, very thick, color=lblue]
                table[x=i, y=erm, col sep=comma] {fig_data/error_quadsensing_200x10_n_iter-10.csv};
            \addplot[mark=square, very thick, color=black]
                table[x=i, y=fix, col sep=comma] {fig_data/error_quadsensing_200x2_n_iter-10.csv};
            \addplot[mark=square, very thick, color=hblue]
                table[x=i, y=fix, col sep=comma] {fig_data/error_quadsensing_200x5_n_iter-10.csv};
            \addplot[mark=square, very thick, color=lblue]
                table[x=i, y=fix, col sep=comma] {fig_data/error_quadsensing_200x10_n_iter-10.csv};
        \end{axis}
    \end{tikzpicture}
    \end{minipage}
    \caption{Performance of spectral initialization for quadratic sensing with
        $X_{\sharp} \in \Rbb^{d \times r}$ in the centralized and distributed
        settings. Here, $d = 100$ (\textbf{left}) and $d = 200$ (\textbf{right}),
        with $n = i \cdot r d$ , $m = 30$. Algorithm~\ref{alg:procrustes-fixing-iterative}
    is invoked with $\texttt{n\_iter} = 10$.}
    \label{fig:quadratic-sensing}
\end{figure}

To that end, we design the following experiment: we set $d \in \set{100, 200}$,
$m = 30$ and vary $r \in \set{2, 5, 10}$. We generate a set of instances with
$X_{\sharp} \sim \Obb_{d, r}$ and let $n$ take on values $n \in
\set{i \cdot r \cdot d, i \in \set{1, \dots, 8}}$ and
apply~\cref{alg:procrustes-fixing-iterative} for better estimates.
In~\cref{fig:quadratic-sensing}, we plot the distance
$\norm{(I_d - X_{\sharp} X_{\sharp}^\T) X_0}_2$ as a function of $i$.
The problem becomes more difficult as the rank $r$ increases;
nevertheless, the distributed initialization scheme weakly recovers $X_{\sharp}$
as long as $n \gtrsim 2rd$ on each machine. In contrast, na{\"i}ve averaging of
the local solutions produces an estimate that is nearly orthogonal to
$X_{\sharp}$ (we omit its depiction).

\section{Discussion}
We presented a communication-efficient algorithm for distributed
subspace estimation, which averages the local estimates in a principled manner and
achieves comparable performance to a purely centralized estimator using the
same amount of data for distributed PCA. Our algorithm generalizes that
of~\cite{GSS17}, which only addresses the rank-$1$ case, and requires a single
round of communication between compute nodes and a central coordinator.

From a theoretical perspective, it is straightforward to adapt our analysis
to the setting where a fixed set of vectors is distributed across machines in
an i.i.d.\ fashion, instead of assuming that all vectors are drawn from the same
distribution; in that case, the objective is to approximate the centralized
\textit{empirical} covariance matrix, which can be viewed as the expectation of
each local empirical covariance matrix (this is in fact the setting depicted
in~\cref{fig:real-pca}).
It may also be possible to leverage the
insight and analysis of~\cite{DS19} to prove improved rates for distributed
covariance estimation with error measured in the $\ell_{2,\infty}$ norm
(which yields entrywise bounds in the rank-$1$ case). Nonetheless, this
will require a more careful analysis of the local expansions of each local
solution, as well as $\ell_{2,\infty}$ concentration bounds for empirical
covariance matrices which are asymptotically tighter than the corresponding
spectral norm bounds.

From a practical perspective, our algorithm is not necessarily confined to
distributed covariance estimation. Instead, Procrustes-fixing local solutions
before averaging can be employed in any estimation problem where local estimates
can be rotated arbitrarily, as we demonstrated with distributed node embeddings
and spectral initialization.

The setting of this paper also gives rise to a few questions to explore in
future work. One example is the following: what if \textit{some of the machines
are compromised} and can return an arbitrary matrix with orthonormal columns
instead of an unbiased estimate of the leading eigenspace?
In this case, we believe
it may be possible to adapt the proposed method, using a robust distance
estimator to choose an appropriate reference solution for~\cref{alg:procrustes-fixing}
with high probability and treating the averaging step as a robust mean estimation
problem.\footnote{This model of failure is an instance of the so-called \textit{Byzantine
generals} problem~\cite{LSP19}.} Finally, it would be interesting to identify
other classes of learning problems, for which our deterministic result can be
used in a black-box fashion to yield competitive estimation rates in the
distributed setting.


\subsection*{Acknowledgements}
We wish to thank the anonymous referees for their careful reading of an earlier
version of the manuscript, as well as their invaluable feedback and suggestions.

\noindent This research was supported by NSF Award DMS-1830274, ARO Award W911NF19-1-0057,
ARO MURI, and JPMorgan Chase \& Co.

\bibliographystyle{hplain}
\bibliography{main}

\appendix
\section{Auxiliary results}
\label{sec:aux-results}
We present a few results used in the proofs of the main text.
\begin{lemma}[Corollary 6.20 in~\cite{Wainwright19}]
    \label{lemma:empirical-covariance-concentration}
    Let $\cD$ be a distribution on $\Rbb^d$ such that $x \sim \cD \Rightarrow
    \norm{x} \leq \sqrt{b}$, and let $\Sigma = \expec[x \sim \cD]{xx^\T}$.
    Then, if $x_1, \dots, x_n \overset{\text{i.i.d.}}{\sim} \cD$, the empirical
    covariance matrix $\widehat{\Sigma} := \frac{1}{n} \sum_{i=1}^n x_i x_i^\T$
    satisfies
    \begin{align}
        \prob{\norm{\widehat{\Sigma} - \Sigma}_2 \geq t}
        & \leq 2 d \expfun{-\frac{n t^2}{2 b(\norm{\Sigma}_2 + t)}}
        \leq 2 d \expfun{-\frac{n}{4} \cdot \min\set{\frac{t^2}{b^2}, \frac{t}{b}}},
        \quad \forall t \geq 0.
    \end{align}
\end{lemma}

\begin{lemma}[Local path independence]
    \label{lemma:local-path-independence}
    Let $A \in \Sbb^{n \times n}$ be a fixed symmetric matrix, and let $\hat{A}$
    be a perturbed version of $A$. Decompose the error $E := \hat{A} - A$ into
    the two following sums:
    \begin{equation}
        E = E_0 + E_1 \equiv E'_0 + E'_1
    \end{equation}
    and define the following perturbed matrices:
    \begin{align}
        \hat{A}_1 &:= A + E_0, \quad \hat{A}_2 := \hat{A}_1 + E_1,
        \quad \tilde{A}_1 := A + E'_0, \quad \tilde{A}_2 := \tilde{A}_1 + E'_1.
    \end{align}
    Let $V \in \Obb_{n, r}$ be the leading invariant subspace of $A$, and construct
    the invariant subspaces $\hat{V}_1, \hat{V}_2$ of $\hat{A}_1, \hat{A}_2$
    so that they satisfy
    \begin{equation}
        \min_{U \in \Obb_r} \norm{\hat{V}_1 U - V}_F = \norm{\hat{V}_1 - V}_F,
        \quad
        \min_{U \in \Obb_r} \norm{\hat{V}_2 U - \hat{V}_1}_F = \norm{\hat{V}_2 - \hat{V}_1}_F.
        \label{eq:near-subspaces}
    \end{equation}
    Moreover, let the invariant subspaces $\tilde{V}_1, \tilde{V}_2$ for
    $\tilde{A}_1, \tilde{A}_2$ be constructed similarly to $\hat{V}_1, \hat{V}_2$
    so that they satisfy~\Cref{eq:near-subspaces}.
    Then, $\hat{V}_2, \tilde{V}_2$ are both principal invariant subspaces for
    $\hat{A} := A + E$, and moreover they satisfy
    \begin{equation}
        \hat{V}_2 = \tilde{V}_2 + T, \quad
        \norm{T}_2 \lesssim \varepsilon^2,
        \quad \varepsilon := \max\set{
            \norm{E_0}_2, \norm{E_1}_2,
            \norm{E'_0}_2, \norm{E'_1}_2
        },
        \label{eq:general-path-independence}
    \end{equation}
    as long as $\varepsilon \leq \frac{\lambda_r(A) - \lambda_{r+1}(A)}{4}$.
\end{lemma}
\begin{proof}
    The proof is essentially outlined in Section 5 of~\cite{Stewart12}. Given
    an invariant subspace $V$ of $A$, the author considers the (unique) basis
    \begin{equation}
        \hat{V}_1 := (V + V_{\perp} P_1) (I + P_1^\T P_1)^{-1/2},
    \end{equation}
    where $V_{\perp}$ is an orthonormal basis for the orthogonal complement of
    $V$, as long as this decomposition of $\hat{V}_1$ exists. However, this is
    precisely the decomposition employed in~\cite[Section 3.2]{DS19}, which is
    guaranteed to exist as long as $\varepsilon$ satisfies the inequality in
    the statement of the Lemma; therein it is shown that $\hat{V}_1$ is the
    (unique) matrix that satisfies
    \begin{equation}
        \min_{U \in \Obb_r} \norm{\hat{V}_1 U - V}_F = \norm{\hat{V}_1 - V}_F.
    \end{equation}
    The rest follows verbatim from the arguments in~\cite[Section 5]{Stewart12}.
\end{proof}

\begin{lemma}[High-probability events]
    \label{lemma:high-prob-events}
    Let $\est^i$ be the local covariance matrices and $X = \expec{xx^\T}$.
    Let $\delta := \lambda_r(X) - \lambda_{r+1}(X) > 0$. Define the following
    events:
    \begin{align}
        \begin{aligned}
        \cE_1 &:= \set{\bignorm{\frac{1}{m} \sum_{i=1}^m \hat{X}^i - X}_2 \leq
        2 \cdot \sqrt{\frac{b^2 \log(2d / p)}{mn}}} \\
            \cE_2 &:= \set{\forall i: \norm{\est^i - X}_2
            \leq \min\set{\frac{\delta}{8}, 2 \cdot \sqrt{\frac{b^2 \log(2dm / p)}{n}}}} \\
        \cE &:= \cE_1 \cap \cE_2,
        \end{aligned}
        \label{eq:good-events}
    \end{align}
    Then, as long as $n \geq 4 \log\frac{2dm}{p}$, we have
    \begin{equation}
        \prob{\cE} \geq 1 - 2p - 2dm \expfun{-\frac{n \delta^2}{4b^2}}.
        \label{eq:prob-E}
    \end{equation}
\end{lemma}
\begin{proof}
First, we rewrite the term in $\cE_1$ as
\begin{align}
    \frac{1}{m} \sum_{i=1}^m \hat{X}^i - \expec{\hat{X}^i} &=
    \frac{1}{mn} \sum_{i=1}^m \sum_{j=1}^n x^{(i)}_{j} {x^{(i)}_{j}}^\T
        - \expec[x \sim \cD]{xx^\T} =
    \frac{1}{mn} \sum_{k=1}^{mn} x_{k} x_{k}^\T - \expec{x_k x_k^\T},
\end{align}
after relabelling, which is the same as the difference between an empirical
covariance matrix of $m \cdot n$ i.i.d. samples and its expectation. Appealing
to~\cref{lemma:empirical-covariance-concentration}, we recover
\begin{align}
    \prob{\bignorm{\frac{1}{m} \sum_{i=1}^m \hat{X}^i - \expec{\hat{X}^i}}_2
    \geq 2 \sqrt{\frac{b^2 \log (2d / p)}{mn}}}
    & \leq 2d \expfun{-\min\set{\log \frac{2 d}{p}, \frac{\sqrt{mn \log \frac{2d}{p}}}{2}}} \notag \\
    & \leq \frac{2dp}{2d} = p,
    \label{eq:centralized-covariance-concentration}
\end{align}
as long as $mn \geq 4 \log \frac{2d}{p}$.

For the terms in $\cE_2$, applying~\cref{lemma:empirical-covariance-concentration}
for each $i$ with $t := \frac{\delta}{8}$ and taking a union bound over all machines
yields
\[
    \prob{\exists i: \norm{\est^i - X}_2 \geq \frac{\delta}{8}} \leq
    2md \expfun{-\frac{n}{4} \min\set{\frac{\delta^2}{b^2}, \frac{\delta}{b}}}
    = 2md \expfun{-\frac{n\delta^2}{4b^2}},
\]
using the fact that $\delta := \lambda_r(X) - \lambda_{r+1}(X)
\leq \lambda_1(X) \leq b$, so that $\min\set{\frac{\delta^2}{b^2}, \frac{\delta}{b}}
= \frac{\delta^2}{b^2}$.
Further, applying~\cref{lemma:empirical-covariance-concentration} for each $i$ with
$u := 2 \sqrt{\frac{b^2 \log(2dm / p)}{n}}$ and a union bound results in
\begin{align*}
    \prob{\exists i: \norm{\est^i - X}_2 \geq u}
    &\leq
    2md \expfun{-\frac{1}{4} \min \set{4 \log (2 dm / p), 2 \sqrt{n \log (2 dm / p)}}} \\
    &\leq
    2md \expfun{-\log (2 dm / p)} = \frac{2mdp}{2 dm} = p,
\end{align*}
whenever $n \geq 4 \log \frac{2dm}{p}$. Combining, we have
\[
    \prob{\exists i: \; \norm{\est^i - X}_2 \geq
    \min\set{\frac{\delta}{8}, 2 \cdot \sqrt{\frac{b^2 \log(2 dm / p)}{n}}}}
    \leq 2dm \expfun{-\frac{n\delta^2}{4b^2}} + p.
\]
Taking a union bound recovers the claimed result.
\end{proof}

\begin{lemma}[Spectral norm of block matrix]
    \label{lemma:block-spectral-norm}
    Consider the matrices $A$, $\tilde{A}$ given by
    \[
        A := \bmx{\bm{a}_1 & \bm{0}_d & \dots & \bm{0}_d \\
                  \bm{0}_d & \bm{a}_2 & \dots & \bm{0}_d \\
                  \vdots & & & \vdots \\
              \bm{0}_d & \bm{0}_d & \dots & \bm{a}_r}
        \in \Rbb^{dr \times r},
        \quad
        \tilde{A} := \bmx{\bm{a}_1 & \dots & \bm{a}_r} \in \Rbb^{d \times r},
    \]
    where $\bm{0}_d$ is the $d$-dimensional all-zeros vector and
    $\bm{a}_{i} \in \Rbb^{d}, \; i \in [r]$. Then, it holds that
    \begin{equation}
        \norm{A}_2 \leq \norm{\tilde{A}^\T}_{2 \to \infty} \leq \norm{\tilde{A}}_2,
        \label{eq:norm-2-2inf-2}
    \end{equation}
    where $\norm{C}_{2 \to \infty} := \max_{i \in [m]} \norm{C_{i, :}}_2$ for a matrix
    $C \in \Rbb^{m \times n}$.
\end{lemma}
\begin{proof}
    We work with the definition of the spectral norm, so that
    \begin{align}
        \norm{A}^2_2 &:= \sup_{x \in \Sbb^{r - 1}} \norm{Ax}_2^2 =
        \sup_{x \in \Sbb^{r - 1}} \norm{\pmx{x_1 \cdot \bm{a}_1^\T & x_2 \cdot \bm{a}_2^\T
        & \dots & x_r \cdot \bm{a}_r^\T}^\T}_2^2 \\
        &=
        \sup_{\set{x \mmid x_1^2 + \dots + x_r^2 = 1}}
        \sum_{i=1}^r \norm{\bm{a}_i}_2^2 x_i^2
        \leq \max_{i \in [r]} \norm{\bm{a}_i}_2^2 =
        \max_{i \in [r]} \norm{\tilde{A}_{:, i}}_2^2 = \norm{\tilde{A}^\T}_{2 \to \infty}^2
    \end{align}
    The second inequality from~\eqref{eq:norm-2-2inf-2} follows from properties
    of the $2 \to \infty$ norm~\cite{CTP19}.
\end{proof}

\section{Omitted proofs}
\label{sec:omitted-proofs}

\subsection{Proof of Lemma~\ref{lemma:local-expansion-1}}
\label{sec:local-expansion-1-proof}
    Using the same steps as in~\cite{DS19}, we arrive at their Eq. (3.7), which
    states that the chosen $\widehat{V}_1$ and $\mtrz$ satisfy
    \begin{align}
        \widehat{V}_1 - V_1 - V_2 \mtrz (I + \mtrz^\T \mtrz)^{-1/2}
        &= V_1 \left((I + \mtrz^\T \mtrz)^{-1/2} - I\right),
        \label{eq:local-expansion-step-1} \\ \Rightarrow
        \norm{\widehat{V}_1 - V_1 - V_2 \mtrz}_2 &\leq
        \norm{V_1 + V_2 \mtrz}_2 \cdot
        \norm{(I + \mtrz^\T \mtrz)^{-1/2} - I}_2,
        \label{eq:local-expansion-step-2}
    \end{align}
    where the first term in the product of the RHS in~\eqref{eq:local-expansion-step-2}
    can be upper bounded via the triangle inequality and submultiplicativity
    of the spectral norm:
    \begin{equation}
        \norm{V_1 + V_2 \mtrz}_2 \leq
        \norm{V_1}_2 + \norm{V_2}_2 \norm{\mtrz}_2 \leq 1 + \norm{\mtrz}_2,
    \end{equation}
    and the second term can be bounded above by $\frac{1}{2} \norm{\mtrz}_2^2$
    as in~\cite{DS19}. Then we simply appeal to~\cite[Lemma 3.4]{DS19}, which
    states that $\norm{\mtrz}_2 \lesssim \frac{\norm{\est - X}_2}{\delta}$ as long
    as the difference $E := \est - X$ satisfies $\norm{E}_2 \leq \frac{\delta}{4}$.
    This is satisfied by~\cref{asm:deterministic}, and combined with the fact
    that $\norm{\mtrz}_2 \leq 1$ concludes the proof.
\qed

\subsection{Proof of Lemma~\ref{lemma:local-expansion-sylvester}}
\label{sec:local-expansion-sylvester-proof}
Recall that we can expand $X = V_1 \Lambda_1 V_1^\T + V_2 \Lambda_2 V_2^\T$,
and rewrite $\hat{X}^i := X + E^i$, which yields the following simplifications
(dropping the superscript for brevity):
\begin{align}
    \begin{aligned}
        \hat{X}_{11} &= V_1^\T (X + E) V_1 = \Lambda_1 + E_{11} &
        \hat{X}_{22} &= V_{2}^\T (X + E) V_2 = \Lambda_2 + E_{22}, \\
    \hat{X}_{12} &= E_{12}, & \hat{X}_{21} &= E_{21}.
    \end{aligned}
    \label{eq:matrix-quadratic-simplify}
\end{align}
Substituting~\cref{eq:matrix-quadratic-simplify} into~\cref{eq:Z-root}
and rearranging, we recover that $\mtrz^{(i)}$ is a solution of the following equation:
\begin{align}
    \mtrz^{(i)} \Lambda_1 - \Lambda_2 \mtrz^{(i)} &=
    \underbrace{-\mtrz^{(i)} E^i_{11} - E^i_{21} + \mtrz^{(i)} E^i_{12} \mtrz^{(i)}
    + E^i_{22} \mtrz^{(i)}}_{\err^{(i)}}.
    \label{eq:diag-sylvester}
\end{align}
Disregarding the fact that $\widehat{Z}$ appears on the right
of~\cref{eq:diag-sylvester}, we see that the latter is a Sylvester equation
of the form $XA - BX = C$, where $A, B$ are diagonal.
Denoting $z_k := \widehat{Z}_{:, k}$ for the $k^{\text{th}}$ column of $\widehat{Z}$
(again dropping the superscripts for simplicity) and expressing the equality
in~\cref{eq:diag-sylvester} column-wise, we arrive at the system of equations:
\begin{align}
    \label{eq:sylvester-sols-system}
    \begin{aligned}
        (\widehat{Z} \Lambda_1)_{:, k} - (\Lambda_2 \widehat{Z})_{:, k} &=
        \err_{:, k}
        \Leftrightarrow
        (\lambda_k I - \Lambda_2) z_k = \err_{:, k} \Leftrightarrow
        z_k = (\lambda_k I - \Lambda_2)^{-1} \err_{:, k},
        \; k \in [r]
    \end{aligned}
\end{align}
since by our assumptions on the eigengap, the matrices appearing on the
left-hand side of~\cref{eq:sylvester-sols-system} are all invertible. Denote
$\bm{\Lambda}_k := (\lambda_k I - \Lambda_2)^{-1} \in \Rbb^{(n - r) \times (n - r)}$,
and gather the above equalities to rewrite (for each machine $i$):
\begin{equation}
    V_2 \mtrz^{(i)} = V_2 \cdot \bmx{
        \bm{\Lambda}_1 & \bm{\Lambda}_2 & \dots &
    \bm{\Lambda}_r}
    \cdot
    \bmx{\err^{(i)}_{:, 1} & 0 & \dots & 0 \\
        0 & \err^{(i)}_{:, 2} & \dots & 0 \\
         \vdots & & & \vdots \\
     0 & 0 & \dots & \err^{(i)}_{:, r}}.
    \label{eq:sylvester-sol-bkdiag}
\end{equation}
To ease notation, denote $\bm{E}_i$ for the block-diagonal matrix on the right-hand side
of~\cref{eq:sylvester-sol-bkdiag}, to arrive at
\begin{align}
    \widehat{Y}_i &= V_2 \bmx{\bm{\Lambda}_1 & \dots & \bm{\Lambda}_r} \cdot
    \bm{E}_i
    \Rightarrow
    \frac{1}{m} \sum_{i=1}^m \widehat{Y}_i =
    V_2 \bmx{\bm{\Lambda}_1 & \dots & \bm{\Lambda}_r} \cdot
    \frac{1}{m} \sum_{i=1}^m \bm{E}_i \\
    \Rightarrow
    \norm{\frac{1}{m} \sum_{i=1}^m \widehat{Y}_i}_2
                  &\leq
                  \max_{i \in [r]} \norm{\bm{\Lambda}_i}_2 \cdot
                  \norm{\frac{1}{m} \sum_{i=1}^m \bm{E}_i}_2
                  \leq \frac{1}{\lambda_r - \lambda_{r+1}}
    \cdot \norm{\frac{1}{m} \sum_{i=1}^m \bm{E}_i}_2,
    \label{eq:sum-bound}
\end{align}
where the last expression is equal to
\begin{equation}
    \sum_{i=1}^m \bm{E}_i =
    \bmx{\sum_{i} \err^{(i)}_{:, 1} & 0 & \dots & 0 \\
         0 & \sum_{i} \err^{(i)}_{:, 2} & \dots & 0 \\
         \vdots & & & \vdots \\
         0 & 0 & \dots & \sum_{i} \err^{(i)}_{:, r}}.
    \label{eq:err-sum-diagonal}
\end{equation}
From~\cref{lemma:block-spectral-norm}, we know that the spectral norm
of the matrix in~\eqref{eq:err-sum-diagonal} can be upper bounded as follows:
\begin{equation}
    \norm{\frac{1}{m} \sum_{i=1}^m \bm{E}_i}_2 \leq
    \norm{\frac{1}{m} \bmx{\sum_{i=1}^m \err^{(i)}_{:, 1} & \dots &
    \sum_{i=1}^m \err^{(i)}_{:, r}}}_2 =
    \norm{\frac{1}{m} \sum_{i=1}^m \err^{(i)}}_2.
    \label{eq:spec-norm-ub}
\end{equation}
Now, notice that
\begin{equation}
    \widehat{E}^{(i)} = - E_{21}^i +
    \underbrace{\widehat{Z}_i E_{12}^i \widehat{Z}_i + E_{22}^i \widehat{Z}_i
        - \widehat{Z}_i E_{11}^i}_{\widehat{R}_i},
    \label{eq:residual-i}
\end{equation}
which we can substitute above and apply the triangle inequality to obtain:
\begin{align}
    \norm{\frac{1}{m} \sum_{i=1}^m \widehat{E}^{(i)}}_2 &\leq
    \norm{V_2^\T \left( \frac{1}{m} \sum_{i=1}^m \hat{X}^i - X \right) V_1}_2
    + \frac{1}{m} \sum_{i=1}^m \norm{\widehat{R}_i}_2 \\
                                                        &\leq
    \norm{\frac{1}{m} \sum_{i=1}^m \hat{X}^i - X}_2
    + \frac{1}{m} \sum_{i=1}^m \norm{\widehat{R}_i}_2.
    \label{eq:Ei-sum}
\end{align}
By~\cite[Lemma 3.4]{DS19}, using the error bound on $\norm{\est^i}_2$
from~\cref{asm:deterministic}, each matrix $\mtrz^{(i)}$ satisfies
\[
    \norm{\mtrz^{(i)}}_2 \lesssim
    \min\set{1, \frac{\norm{\err^i}_2}{\delta}}
    \]
and so the quantities defined in~\cref{eq:residual-i} satisfy
\begin{align*}
    \max\set{ \norm{\mtrz^{(i)} \err^i_{12} \mtrz^{(i)}}_2,
        \norm{\mtrz^{(i)} \err^i_{11}}_2,
    \norm{\err^i_{22} \mtrz^{(i)}}_2}
    & \lesssim \frac{\norm{\err^i}_2^2}{\delta}, \quad \forall i.
\end{align*}
Plugging the above bounds back into~\cref{eq:Ei-sum} and then
into~\cref{eq:spec-norm-ub}, and combining with~\cref{eq:sum-bound}, we
conclude that
\begin{equation}
    \norm{\frac{1}{m} \sum_{i=1}^m \mtry^{(i)}}_2
    \lesssim \frac{1}{\delta^2 m} \sum_{i = 1}^m \norm{\est^i - X}_2^2
    + \frac{1}{\delta} \norm{\frac{1}{m} \sum_{i=1}^m \est^i - X}_2.
\end{equation}
\qed

\subsection{Proof of~\cref{lemma:path-independence}}
\label{sec:path-independence-proof}
    We treat the matrix $\est^i$ as a perturbation of $X$ in the two ways
    outlined below:
    \begin{align}
        \est^i := X + \underbrace{(\est^i - X)}_{=: E_0} +
            \underbrace{0}_{=: E_1}, \quad
            \est^i := X + \underbrace{\est^1 - X}_{=: E'_0}
            + \underbrace{\est^i - \est^1}_{=: E'_1}
    \end{align}
    We intend to apply Lemma~\ref{lemma:local-path-independence}; to that end,
    notice that by~\cref{asm:deterministic}, all the conditions on the
    magnitude of the local perturbations are satisfied for any $i$, since
    \begin{equation}
        \norm{\est^i - \est^1}_2 \leq \norm{\est^i - X}_2 + \norm{X - \est^1}_2
        \leq \frac{\delta}{4},
    \end{equation}
    while the other conditions are immediate.

    Note that $\mtrv_1^{(i)}$, the eigenvector matrix outlined in
    \cref{proposition:average-subspace}, has been chosen to satisfy
    \eqref{eq:nearest-matrix} and thus is identical to the eigenvector matrix
    $\hat{V}_1$ from Lemma~\ref{lemma:local-path-independence}. In turn, since
    $E_1 = 0$, $\hat{V}_1 = \hat{V}_2 \equiv \mtrv_1^{(i)}$ in the notation of
    that Lemma.

    For the second decomposition, we also note that
    $A_1 := X + (\est^1 - X) = \est^1$ has $\mtrv_1^{(1)}$ as its leading
    invariant subspace satisfying~\eqref{eq:nearest-matrix} which in turn has been
    chosen in a way that lets us identify it with $\tilde{V}_1$ of
    Lemma~\ref{lemma:local-path-independence}. Moreover, since $\tilde{V}_2$
    is aligned with $\tilde{V}_1$ in Lemma~\ref{lemma:local-path-independence},
    and we have identified $\tilde{V}_1 = \mtrv_1^{(1)}$, $\tilde{V}_2$
    can be identified with $\tilde{V}^{(i)}$, the Procrustes-fixed estimate,
    as the latter has also been explicitly aligned with $\mtrv_1^{(1)}$.
    By the result of Lemma~\ref{lemma:local-path-independence}, we deduce that
    \begin{equation}
        \tilde{V}^{(i)} = \mtrv_1^{(i)} + T^{(i)},
        \quad \norm{T^{(i)}}_2 \lesssim \varepsilon^2,
    \end{equation}
    with $\varepsilon := \max\set{\norm{\est^i - X}_2, \norm{\est^i - \est^1}_2,
    \norm{\est^1 - X}_2}$. Using $(a + b)^2 \leq 2(a^2 + b^2)$, we can further
    bound
    \begin{equation}
        \norm{\est^i - \est^1}^2_2 \leq 2 \left( \norm{\est^i - X}^2_2 +
        \norm{\est^1 - X}^2_2 \right),
    \end{equation}
    which leads to the claimed inequality for $\norm{T^{(i)}}_2$.
\qed

\subsection{Proof of~\cref{theorem:stablerank-rate}}
\label{sec:stablerank-rate-proof}

The proof is essentially the same as the proof of~\cref{theorem:procustes-fixing-works},
with a slight modification of the high probability events, which we provide for
completeness. Without loss of generality, we may rescale so that $\norm{X}_2 = 1$.
When $\cD$ is a subgaussian distribution, the following concentration
inequality, appearing as~\cite[Exercise 9.2.5]{Vershynin18}, is applicable
(with $r_{\star} := \textsf{intdim}(X)$):
\begin{equation}
    \prob{\norm{\hat{X}^i - X}_2 \geq
    C \left( \sqrt{\frac{r_{\star} + u}{n}} + \frac{r_{\star} + u}{n}\right)}
    \leq 2\expfun{-u}.
    \label{eq:subg-cov-estimation}
\end{equation}
Applying~\eqref{eq:subg-cov-estimation} with $u := \log(m / p)$ and taking a
union bound over $m$ machines yields
\begin{equation}
    \prob{\exists i: \norm{\hat{X}^i - X}_2 \geq
        C \left( \sqrt{\frac{r_{\star} + \log(m / p)}{n}} +
    \frac{r_{\star} + \log(m / p)}{n} \right)} \leq p
\end{equation}
In particular, as long as $n \gtrsim \frac{r_{\star} + \log(m/p)}{\delta^2}$,
we also satisfy the condition $\max_{i} \norm{X^i - X}_2 \leq \frac{\delta}{8}$
with the same probability. \Cref{eq:subg-cov-estimation} will also yield
\begin{equation}
    \prob{\bignorm{\frac{1}{m} \sum_{i=1}^m \hat{X}^i - X}_2 \geq C
        \left(\sqrt{\frac{r_{\star} + c_1 \log n}{mn}} +
    \frac{r_{\star} + c_1 \log n}{mn}\right)} \leq \frac{2}{n^{c_1}},
\end{equation}
where $c_1$ can now be adjusted to achieve the desired probability at small
extra cost. Finally, if $n \gtrsim \max\set{1, \frac{1}{\delta^2}} \cdot
(r_{\star} + \log(m / p))$, and since $mn \gtrsim (r_{\star} + c_1 \log n)$,
it is easy to check that the terms in square roots above dominate, and the
high probability events of~\eqref{eq:good-events} become
\begin{align}
    \begin{aligned}
    \cE_1 &:= \set{\bignorm{\frac{1}{m} \sum_{i=1}^m \hat{X}^i - X}_2 \leq C
        \sqrt{\frac{r_{\star} + c_1 \log n}{mn}}}, \\
    \cE_2 &:= \set{\forall i: \norm{\est^i - X}_2
    \leq \min\set{\frac{\delta}{8}, \sqrt{\frac{r_{\star} + \log(m /p)}{n}}}}, \\
    \end{aligned}
    \label{eq:good-events-mod}
\end{align}
and hold with probability at least $1 - p - 2n^{-c_1}$. The rest of the proof
follows the same steps using the above events and probabilities instead
of~\cref{lemma:high-prob-events}.
\qed

\end{document}